\documentclass{article}

 \usepackage[main, final]{neurips_2026}

\usepackage[utf8]{inputenc} 
\usepackage[T1]{fontenc}    
\usepackage{hyperref}       
\usepackage{url}            
\usepackage{booktabs}       
\usepackage{amsfonts}       
\usepackage{nicefrac}       
\usepackage{microtype}      
\usepackage{xcolor}         

\usepackage{graphicx}
\usepackage{amsmath} 
\usepackage{array} 
\usepackage[table]{xcolor} 
\usepackage{caption} 

\usepackage{algorithm} 
\usepackage{algpseudocode} 
\usepackage{amssymb} 
\usepackage{wrapfig} 
\usepackage{multirow} 
\usepackage{makecell} 
\usepackage{graphicx} 

\usepackage{subcaption} 
\usepackage{arydshln} 
\usepackage[table]{xcolor} 
\usepackage{enumitem} 

\algrenewcommand\algorithmicrequire{\textbf{Input:}}
\algrenewcommand\algorithmicensure{\textbf{Output:}}
\usepackage{algpseudocode}

\definecolor{customgreen}{HTML}{89BD6D} 
\definecolor{customorange}{HTML}{D79B00} 
\definecolor{customred}{HTML}{AB4E4B} 
\definecolor{customblue}{HTML}{6C8EBF} 

\title{Self-Evolution Reasoner: Continuous Optimization via Policy-Intrinsic Exploration and Exploitation}

\author{%
  \begin{tabular}{c}
  \bf{Wenhang Shi\textsuperscript{1} \quad 
  Yiren Chen\textsuperscript{2} \quad 
  Zhe Zhao\textsuperscript{3} \quad 
  Jinhao Dong\textsuperscript{1} \quad
  Wei Lu\textsuperscript{1} \quad
  Xiaoyong Du\textsuperscript{1}\thanks{Corresponding author}
  } \\
  \\[-2.5ex] 
  \begin{tabular}{c}
  \normalfont\textsuperscript{1}Renmin University of China \\
  \normalfont\textsuperscript{2}Peking University \\
  \normalfont\textsuperscript{3}Tencent
  \end{tabular} \\
  \\[-2.5ex] 
  \texttt{\{wenhangshi, dongjinhao, lu-wei, duyong\}@ruc.edu.cn,} \\ \texttt{yrchen92@pku.edu.cn, nlpzhezhao@tencent.com}
  \end{tabular}
}

\begin{document}

\maketitle

\begin{abstract}
Reinforcement learning with verifiable rewards (RLVR) enables large language models (LLMs) to continuously self-improve their reasoning capabilities.
Compared to RLVR's scalar rewards, recent self-supervised on-policy distillation methods leverage denser supervision signals.
However, both paradigms suffer from exploration bottlenecks,  struggling to escape local optima when the policy fails to sample valid solutions.
Relying on external demonstrations can bypass this issue; however, the off-policy nature hinders genuine self-improvement and the mismatched-distribution reasoning can not be fully exploited.
In this work, we propose Self-Evolution Policy Optimization (SEPO), a framework that drives continuous reasoning capability evolution via guided exploration and in-distribution exploitation.
Specifically, SEPO utilizes labels to elicit correct reasoning from the policy, unlocking the exploration potential.
By treating current policy conditioned on this in-distribution reasoning as a self-teacher, SEPO distills the newly discovered solution back into the policy.
Through this iterative cycle of policy-intrinsic exploration and exploitation, SEPO achieves robust and continuous capability evolution.
Across scientific, tool-use, and complex logical reasoning domains, SEPO consistently achieves the best performance under identical training time or steps, demonstrating superior efficiency and convergence ceilings. Notably, SEPO maintains stable improvements even in scenarios where baselines severely underperform or fail entirely, such as on complex tasks or with small-scale models. Our code is available at \url{https://github.com/Eric8932/SEPO}.


\end{abstract}

\section{Introduction}
Reinforcement learning with verifiable rewards (RLVR) enables large language models (LLMs) to develop sophisticated reasoning capabilities by rewarding their own correct rollouts~\cite{DBLP:journals/corr/abs-2402-03300,DBLP:journals/corr/abs-2501-12948,DBLP:journals/corr/abs-2505-09388}.
Recently, on-policy self-distillation methods have emerged, leveraging denser supervision signals to achieve more efficient self-improvement~\cite{lu2025onpolicy,DBLP:journals/corr/abs-2601-20802}. 
They treat the current policy with its own correct rollouts in the context as a self-teacher, distilling the enriched predictions back into the policy, thoroughly exploiting the successful solutions.
Although both paradigms can continuously improve reasoning capabilities, their optimization relies entirely on the policy's unguided trials, making them susceptible to exploration bottlenecks.
When the policy fails to sample a valid solution, these methods cannot formulate effective supervision signals, causing the optimization to stagnate.

Recent works introduce external solutions from experts or proprietary models to mitigate the exploration bottleneck, which generally fall into two categories.
The first category distills expert demonstrations for imitation learning, either by fine-tuning on them as an auxiliary training objective~\cite{DBLP:journals/corr/abs-2508-11408,DBLP:journals/corr/abs-2506-19767}, or by placing them within the student's context to construct an expert-imitated distillation teacher ~\cite{DBLP:journals/corr/abs-2601-19897, DBLP:journals/corr/abs-2601-18734,DBLP:journals/corr/abs-2603-11178}. 
The second category integrates expert data to guide RL exploration. They either incorporate it into the same GRPO group with the on-policy rollouts for joint learning~\cite{DBLP:journals/corr/abs-2506-19767,DBLP:journals/corr/abs-2504-14945}, or extract prefixes to serve as initial contexts, making it easier to sample correct solutions~\cite{DBLP:journals/corr/abs-2512-17636,DBLP:journals/corr/abs-2601-18779}.
While relying on external experts circumvents exploration failures, it introduces severe off-policy challenges and causes policy's own exploration to stagnate.
Furthermore, the distribution mismatches prevent the policy from fully internalizing these reasoning patterns.
Ultimately, the dual deficits in exploration and exploitation stifle continuous improvement and impose a strict performance ceiling.

In this work, we propose Self-Evolution Policy Optimization (SEPO), a framework that drives continuous capability evolution via guided exploration and in-distribution exploitation.
Specifically, SEPO utilizes ground-truth labels as privileged hints to elicit correct reasoning from the policy itself.
This alleviates the exploration bottleneck by shifting the learning threshold from generating solutions from scratch to performing guided deduction toward correct answers.
By treating the current policy conditioned on this in-distribution reasoning as a teacher, SEPO distills the context-enriched predictions back into the standard policy, ensuring thorough internalization of the newly discovered  reasoning.
Ultimately, through this iterative cycle of policy-intrinsic exploration and exploitation, SEPO achieves a more sustained capability evolution and a higher performance ceiling.


Across extensive experiments on scientific, tool-use, and logical reasoning tasks, SEPO consistently outperforms strong RLVR and self-distillation methods.
It not only demonstrates the best convergence performance, indicating a higher ceiling, but also exhibits superior training efficiency, surpassing baseline performance in significantly less time or steps. Notably, even in scenarios where baselines underperform or fail entirely, such as in highly complex logical reasoning tasks or when scaling down to weak models, SEPO maintains stable and continuous performance improvements.
Our contributions are summarized as follows:
\begin{itemize}[topsep=-1pt, itemsep=-2.5pt, leftmargin=0.3cm]
\item We propose SEPO, an on-policy self-distillation framework that drives continuous reasoning evolution through superior exploration and exploitation. 
By utilizing label to guide reasoning exploration and constructing in-distribution teacher to exploiting it, SEPO iteratively self-improves.

\item We present a theoretical analysis revealing how exploration and exploitation deficits limit the optimization of existing methods. Crucially, our analysis formally validates that SEPO fundamentally circumvents these bottlenecks, guaranteeing sustained and effective self-evolution.

\item SEPO demonstrates superior training efficiency and final performance across diverse reasoning tasks. Even in scenarios where baselines fail, SEPO robustly secures continuous improvements. 
\end{itemize}


\section{Background}
\subsection{Reinforcement learning with verifiable rewards}
Reinforcement learning with verifiable rewards (RLVR) improves the reasoning ability of large language models by optimizing them against verifiable outcome-level rewards~\cite{DBLP:journals/corr/SchulmanWDRK17,DBLP:journals/corr/abs-2507-20534}.
Given an input problem $x$ with ground-truth label $y^\ast$, a policy $\pi_\theta$ generates a solution $\tau = (z, \hat{y})$, where $z$ denotes the intermediate reasoning process and $\hat{y}$ the final prediction, and receives a scalar reward $r(\tau; y^\ast) = \mathbb{I}[\hat{y} = y^\ast]$, which is typically binary in reasoning tasks.
Recently, RLVR is predominantly implemented via Group Relative Policy Optimization (GRPO)~\cite{DBLP:journals/corr/abs-2402-03300}. 
For each problem $x$, the policy samples a group of $G$ solutions $\{\tau_i\}_{i=1}^G$. The outcome reward $r_i$ for each solution is converted into a group-relative advantage, e.g., $A^{GPRO}_{i,t} = \frac{r_i - \mathrm{mean}(\{r_j\}_{j=1}^G)}{\mathrm{std}(\{r_j\}_{j=1}^G) + \epsilon}$. Crucially, this sequence-level advantage is applied uniformly to every token $t$ within the $i$-th solution.


Despite its broad application in enhancing reasoning, RLVR suffers from two fundamental limitations. First, its supervision is sparse and purely outcome-level: the uniform advantage $A_{i,t}$ provides no direct credit assignment for intermediate reasoning steps~\cite{DBLP:journals/corr/abs-2502-01456,DBLP:journals/tmlr/KhalifaALKPLLW26}.
Second, RLVR relies on the policy possessing robust exploration ability. If the policy fails to sample any correct solutions within a group, the algorithm generates no effective learning signal, causing optimization to stagnate~\cite{DBLP:journals/corr/abs-2510-19807,DBLP:journals/corr/abs-2510-09388}.



\subsection{On-policy context distillation}
Recently, several on-policy context distillation (OPCD) methods leverage denser supervision signals to achieve reasoning self-improvement~\cite{DBLP:journals/corr/abs-2601-20802,DBLP:journals/corr/abs-2601-19897,DBLP:journals/corr/abs-2601-18734}. 
OPCD conditions the student policy $\pi_\theta(\cdot \mid x)$ on  a privileged context $c$ to construct a stronger self-teacher $\pi_{\text{tea}}(\cdot \mid x, c)$.
This context $c$ typically comprises a valid reasoning path, sourced either from student's successful rollouts or external experts, as a reference to simplify the problem-solving process.
Specifically, at each training step, solutions $\tau_i \sim \pi_\theta(\cdot \mid x)$ are sampled from the unhinted student policy. 
In contrast to GRPO's static sequence-level advantage, OPCD provides a token-specific supervision through on-policy distillation~\cite{DBLP:conf/iclr/Gu0WH24,lu2025onpolicy}.
At each token $t$ within solution $i$, the student assimilates the teacher's predictions by minimizing the Kullback-Leibler (KL) divergence over the student's own generated prefix $\tau_{i,<t}$: $ A^{\mathrm{OPCD}}_{i,t} = \mathbb{D}_{\mathrm{KL}} \left( \pi_{\text{tea}}(\cdot \mid x, c, \tau_{i,<t}) \,\|\, \pi_\theta(\cdot \mid x, \tau_{i,<t}) \right)$~\cite{kullback1951information,DBLP:journals/corr/HintonVD15}. 
Since the teacher shares the student's evolving weights, its contextual reasoning capabilities progressively improve throughout a iterative process, creating a continuous self-improvement flywheel.


\subsection{RLVR and OPCD}\label{subsec_rlvr_and_opcd}
Fundamentally, OPCD can be viewed as a proxy objective for RLVR~\cite{DBLP:journals/corr/abs-2601-19897}. From a variational perspective, the optimal policy $\pi^*$ that maximizes expected reward satisfies $\pi^*(\tau \mid x) \propto \pi_{\text{ref}}(\tau \mid x)\exp(r(\tau)/\eta)$, where $\eta$ is a temperature parameter~\cite{DBLP:conf/emnlp/KorbakPB22,rafailov2023direct}. 
RLVR methods such as GRPO move toward $\pi^*$ using sparse rewards, whereas OPCD instead constructs a teacher policy $\pi_{\text{tea}}(\cdot \mid x, c)$ as an approximation to $\pi^*(\cdot \mid x)$ and minimizes its distance to the student policy (with detailed derivations in Appendix~\ref{app:subsec_variation}). 
Therefore, when the teacher is a good proxy for $\pi^*$, context distillation can drive iterative capability improvement similarly to RLVR, but with much denser supervision. Its effectiveness and stability, however, critically depend on the choice of privileged context $c$, which determines the quality and compatibility of the teacher distribution.

Existing OPCD methods have critical issues when constructing the context. Relying on student's direct rollout ($c = \tau_{\text{self}}$) guarantees on-policy learning. However, its existence is bounded by the marginal probability of the model's exploration success, $P_{\text{self}}(x) = \mathbb{E}_{\tau \sim \pi_\theta}[\mathbb{I}(r(\tau)=1)]$. If the policy fails to solve a complex problem within a rollout group, the empirical estimate $\hat{P}_{\text{self}}(x) \to 0$, rendering the teacher undefined and halting the distillation process.
Alternatively, utilizing solutions from external experts ($c = \tau_{\text{expert}}$) bypasses the exploration bottleneck. However, these off-policy contexts cannot co-evolve with the student's optimization, inherently imposing a ceiling on self-improvement.
Even worse, this paradigm introduces a severe distribution shift, as the teacher $\pi_{\text{tea}}$ often occupies a reasoning subspace outside the student's natural support ($\text{supp}(\pi_{\text{tea}}) \not\subseteq \text{supp}(\pi_\theta)$)~\cite{kumar2019stabilizing,levine2020offline}. 
Minimizing this out-of-distribution divergence forces drastic, localized behavioral shifts rather than naturally internalizing the expert's reasoning patterns.
This insufficient exploitation prevents the policy from  approaching the ceiling set by experts. Full theoretical analyses are in Appendix~\ref{app:subsec_theory_limit_opcd}.

\section{Method: Self-Evolution Policy Optimization}
\begin{figure*}[t]
\centering
\includegraphics[scale = 0.445]{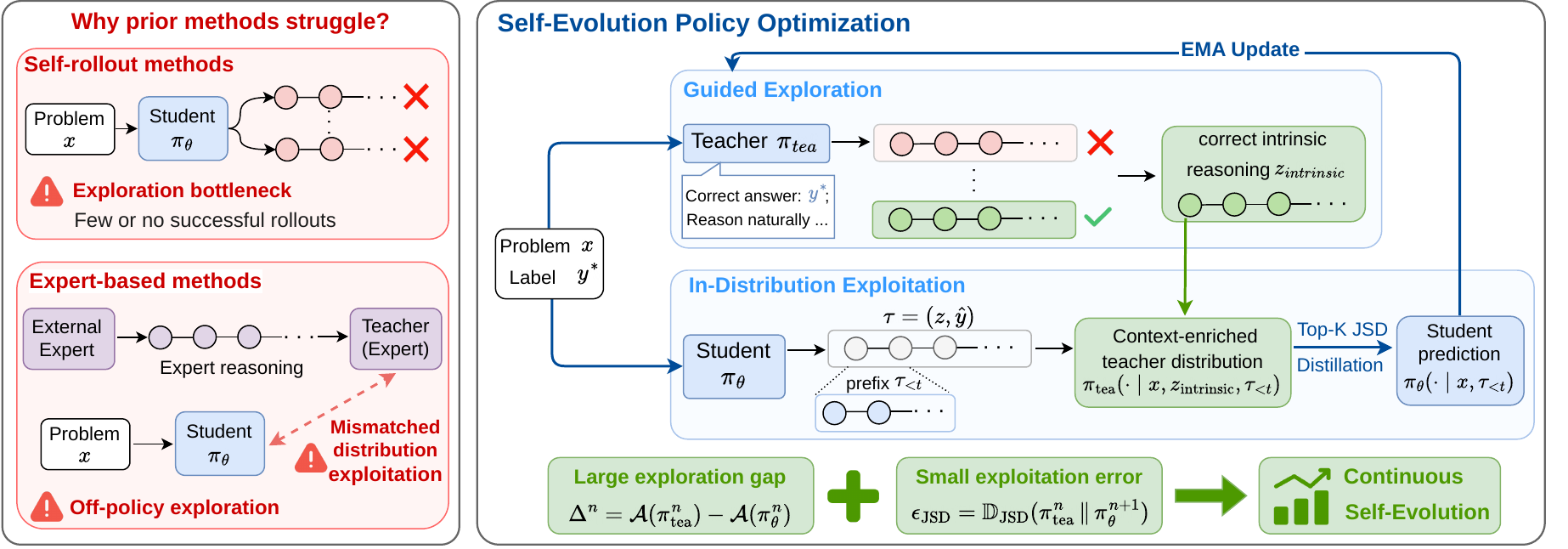}
\vspace{-0.2cm}
\caption{Overview of Self-Evolution Policy Optimization (SEPO) framework. 
SEPO couples guided exploration with in-distribution exploitation, overcoming the exploration bottlenecks and distribution-mismatched exploitation of prior methods, enabling more sustained self-evolution.}
\vspace{-0.2cm}
\label{fig_method}
\end{figure*}

\paragraph{Motivation}
As established, constructing an effective proxy for the optimal policy $\pi^*$ requires a privileged context $c$ that is simultaneously reliably accessible (to bypass the exploration bottleneck) and strictly on-policy (to avoid distribution shifts).
To achieve this, as illustrated in Figure~\ref{fig_method}, we develop the Self-Evolution Policy Optimization (SEPO) framework.
By harnessing a synergy between guided exploration and in-distribution exploitation, SEPO iteratively mines and elevates policy-intrinsic reasoning capacity, realizing  continuous self-evolution.

\begin{wrapfigure}{r}{0.54\linewidth}
\vspace*{-1.3cm}
\refstepcounter{algorithm}
\hrule
\vspace{0.3em}
{\bfseries Algorithm \thealgorithm\ \, SEPO Framework\label{alg:sepo}\par} 
\vspace{0.3em}
\hrule

\begin{algorithmic}[1]
\Require Language Model $\pi_{\theta}$, initial weights $\theta_0$, Dataset $\mathcal{D}$, total steps $N$

\State Initialize $\theta\leftarrow\theta_0$ and $\theta_{\mathrm{tea}}\leftarrow\theta_0$

\For{$n = 0$ to $N-1$}
    \State Sample a batch $\{(x_i, y_i^*)\}_{i=1}^B \sim \mathcal{D}$
    \For{each $(x, y^*)$ in the batch}
        \State Sample rollouts $\tau \sim \pi_\theta(\cdot \mid x)$
        \State Elicit reasoning $z_{\mathrm{in}} \sim \pi_{\theta_{\mathrm{tea}}}(\cdot \mid x, y^*)$
    \EndFor
    \State Update $\theta$ by minimizing Eq.~\ref{equa_loss_sepo}
    \State $\theta_{\mathrm{tea}}\leftarrow(1-\beta)\theta_{\mathrm{tea}}+\beta\theta$

\EndFor

\State \Return $\pi_{\theta}$
\end{algorithmic}
\vspace{0.3em}
\hrule
\vspace*{-0.5cm}
\end{wrapfigure}

\subsection{The SEPO framework}
To circumvent the exploration bottleneck while preserving on-policy exploitation, the Self-Evolution Policy Optimization (SEPO) framework integrates guided reasoning exploration and in-distribution exploitation processes into a synchronized online training loop.
As shown in Algorithm~\ref{alg:sepo}, at each training step, given a problem $x$ and its ground-truth label $y^*$, the student policy $\pi_\theta$ first samples rollouts $\tau \sim \pi_\theta(\cdot \mid x)$. Concurrently, the teacher policy $\pi_{\text{tea}}$ receives a label-informed prompt (e.g., "Correct answer: $y^*$; Reason naturally ...") to elicit a set of reasoning processes and retain only the correct ones, $z_{\text{intrinsic}} \sim \pi_{\text{tea}}(\cdot \mid x, y^*)$. 
Conditioned on this privileged context $z_{\text{intrinsic}}$, the stop-gradient teacher computes the next-token distribution over student's rollout prefix $\tau_{<t}$. 
By minimizing the divergence between the student's predictions and the context-informed teacher's distribution, SEPO continuously pulls the student's unhinted reasoning toward its own label-guided potential.
The detailed training procedure is provided in Algorithm~\ref{alg:sepo_complete}.

\paragraph{Teacher initialization and EMA update.} 
SEPO initializes $\pi_{\text{tea}}$ with the student's weights $\theta$ and updates it via an Exponential Moving Average (EMA): $\theta_{\text{tea}}\leftarrow(1-\beta)\theta_{\text{tea}}+\beta\theta$, where $\beta \in [0, 1)$.
This momentum-based strategy serves two purposes: it ensures the context $z_{\text{intrinsic}}$ evolves synchronously with the student's growing capabilities to facilitate continuous self-improvement, and it provides a smoothed, implicitly ensembled target that prevents confirmation bias and stabilizes learning.

\paragraph{Stable distillation via bounded divergence and tail truncation.}
Standard distillation typically relies on Forward KL divergence ($\mathbb{D}_{\text{KL}}(\pi_{\text{tea}} \,\|\, \pi_\theta)$), which enforces mode-covering behavior~\cite{minka2005divergence,DBLP:journals/corr/HintonVD15}. It heavily penalizes the student for assigning low probability to preferred actions, frequently triggering gradient explosions and catastrophic forgetting~\cite{DBLP:journals/corr/abs-2510-18874}. 
In contrast, Reverse KL ($\mathbb{D}_{\text{KL}}(\pi_\theta \,\|\, \pi_{\text{tea}})$) induces mode-seeking behavior~\cite{DBLP:conf/iclr/Gu0WH24}. While this protects the model's base capabilities by only penalizing student's generations outside the teacher's support, it strictly stifles exploration, causing the student to collapse into  rigid reasoning paths.
To mitigate these extremes, SEPO adopts the Jensen-Shannon Divergence (JSD)~\cite{DBLP:journals/tit/Lin91}. As a symmetric and bounded metric, JSD naturally balances mode-covering and mode-seeking behaviors, providing stable gradients that allow the student to safely assimilate the elicited reasoning.
Furthermore, we approximate the analytic KL divergence by restricting the learning distribution to the top-$K$ tokens predicted by the student model~\cite{DBLP:journals/corr/abs-2601-20802}. This tail truncation significantly reduces the memory footprint while preserving the performance.
Integrating these components, the unified optimization objective for SEPO is formulated as:

\vspace{-0.3cm}
\begin{equation}
    \mathcal{L}_{\mathrm{SEPO}}(\theta) = \mathbb{E}_{x \sim \mathcal{D}}\left[ \mathbb{E}_{\tau \sim \pi_\theta} \left[ \frac{1}{|\tau|} \sum_{t=1}^{|\tau|} \mathbb{D}_{\mathrm{JSD}}^{\operatorname{Top}_K} \left( \tilde{\pi}_{\text{tea}}(\cdot \mid x, z_{\text{intrinsic}}, \tau_{<t}) \,\|\, \pi_\theta(\cdot \mid x, \tau_{<t}) \right) \right] \right].
\label{equa_loss_sepo}
\end{equation}

\subsection{Theoretical analysis of SEPO}
\paragraph{SEPO achieves continuous self-improvement.} 
To demonstrate how SEPO achieves continuous self-improvement, we analyze its training dynamics using a generalized accuracy lower bound~\cite{DBLP:conf/icml/KakadeL02,DBLP:conf/icml/SchulmanLAJM15}.
Let $\mathcal{A}(\pi)$ denote the expected accuracy of a policy. During exploration of privileged context, SEPO leverages label $y^*$ to sample from the posterior reasoning distribution $P(z \mid x, y^*)$ rather than the unguided prior $P(z \mid x)$.
This posterior conditioning exponentially prunes the vast search space, yielding a clear capability advantage over the unhinted student at any training step $n$.
We formalize this advantage as the Exploration Gap: $\Delta^{n} = \mathcal{A}(\pi_{\text{tea}}^{n}) - \mathcal{A}(\pi_\theta^{n})$.
During distillation, SEPO optimizes the student to minimize the Jensen-Shannon Divergence. Let $\epsilon_{\text{JSD}} = \mathbb{D}_{\text{JSD}}(\pi_{\text{tea}}^{n} \,\|\, \pi_\theta^{n+1})$ denote the residual exploitation error, which quantifies the inevitable optimization imperfection remaining after step $n$.
By relating JSD to the Total Variation distance via information-theoretic bounds, this residual error induces a worst-case performance penalty of $\mathcal{O}(\sqrt{\epsilon_{\text{JSD}}})$. 
Consequently, the expected accuracy of the updated student policy at step $n+1$ is  bounded by:
$$\mathcal{A}(\pi_\theta^{n+1}) \ge \mathcal{A}(\pi_{\text{tea}}^{n}) - \mathcal{O}(\sqrt{\epsilon_{\text{JSD}}}) \implies \mathcal{A}(\pi_\theta^{n+1}) \ge \mathcal{A}(\pi_\theta^{n}) + \Delta^{n} - \mathcal{O}(\sqrt{\epsilon_{\text{JSD}}}).$$
As long as the exploration gap outweighs the exploitation error ($\Delta^{n} > \mathcal{O}(\sqrt{\epsilon_{\text{JSD}}})$), the student policy improves.
Through SEPO's synchronous evolution, teacher's capability elevates alongside the student, continuously regenerating a substantial $\Delta^{n}$ to drive continuous self-improvement. 
Ultimately, as the student's intrinsic capabilities grow, the posterior advantage of the label diminishes. Training converges to its empirical upper bound when the exploration gap shrinks to the level of the exploitation error ($\Delta^{n} \approx \mathcal{O}(\sqrt{\epsilon_{\text{JSD}}})$). Detailed theoretical analysis is in Appendix~\ref{app:subsec_theory_SEPO}.

\paragraph{SEPO's superiority over existing methods.}
The recursive inequality explains the superiority of SEPO over other OPCD methods, as SEPO uniquely optimizes both critical terms to maximize the performance upper bound:
1) For methods using student's own rollouts as context, $\Delta$ is bottlenecked by the low prior probability of directly sampling a valid solution. Consequently, their performance ceiling is prematurely locked at the model's unguided exploration limit.
In contrast, SEPO guarantees a more persistently positive $\Delta$. By utilizing guided exploration, SEPO raises the upper bound from the autonomous discovery limit to the posterior exploration limit—the true extent of what the model can solve when accurately guided by the outcome.
2) Methods using external solutions as context may initially provide a high teacher accuracy, but they introduce severe support mismatch between the expert's latent manifold and the student's weights. This mismatch makes the exploitation error $\epsilon_{\text{JSD}}$ intrinsically enormous. When the exploitation error exceeds the exploration gap, the performance gain vanishes, stagnating the optimization.
In contrast, SEPO's strictly on-policy exploration guarantees a minimal $\epsilon_{\text{JSD}}$, allowing the student to absorb the teacher's capability with little optimization friction.


\section{Experiments}

\subsection{Experimental settings}\label{subsec_settings}
We evaluate on reasoning tasks spanning diverse difficulties and paradigms (knowledge, planning, and logic), which are widely employed as standard testbeds to demonstrate method efficacy~\cite{DBLP:journals/corr/abs-2601-20802,DBLP:journals/corr/abs-2601-19897,DBLP:journals/corr/abs-2601-18734}:
\begin{itemize}[topsep=-1pt, itemsep=-2.5pt, leftmargin=0.3cm]
\item \textbf{Science Q\&A}: Undergraduate-level scientific reasoning tasks across Chemistry, Physics, Biology, and Material domains, using the L3 subsets from SciKnowEval~\cite{DBLP:journals/corr/abs-2406-09098}.
\item \textbf{Tool Use}: Planning tool invocations based on user queries and provided tool specifications, utilizing the ToolAlpaca dataset~\cite{DBLP:journals/corr/abs-2306-05301}.
\item \textbf{Logic}: High-difficulty logical reasoning tasks, evaluated on two subsets (Arrow\_Maze and Dyck\_Language) from SynLogic~\cite{DBLP:journals/corr/abs-2505-19641}.
\end{itemize}
We apply a train-test split for each dataset to assess in-domain generalization. We employ models with diverse architectures, specifically Qwen3-8B~\cite{DBLP:journals/corr/abs-2505-09388} and Olmo3-7B-Instruct~\cite{DBLP:journals/corr/abs-2512-13961}, as well as Qwen3 models of varying scales. To ensure robust evaluation, all experimental results are averaged across three independent runs with different random seeds. We report the peak test accuracy (avg@16), measured as the average over 16 sampled responses, achieved within a fixed time or training step.

\paragraph{Baselines}
We evaluate SEPO against GRPO and other OPCD methods.
We implement an enhanced GRPO baseline following~\cite{DBLP:journals/corr/abs-2601-20802}, integrating several recent algorithmic improvements, such as asymmetric clipping~\cite{DBLP:journals/corr/abs-2503-14476} and off-policy data correction~\cite{yao2025your}. 
We evaluate both on-policy and off-policy variants of GRPO, and primarily report the results of the better-performing off-policy version.
For the OPCD baselines, we compare three distinct teacher contexts: OPCD (\textit{Self}) uses student's own correct rollouts~\cite{DBLP:journals/corr/abs-2601-20802}; OPCD (\textit{Expert}) utilizes reasoning processes from DeepSeek-V3.2, which yields the highest performance among several candidate experts~\cite{DBLP:journals/corr/abs-2601-19897,DBLP:journals/corr/abs-2601-18734, DBLP:journals/corr/abs-2512-02556}; and additionally, we include OPCD (\textit{Label}), a simplified variant of SEPO that solely provides the ground-truth label within the context.
To ensure a fair comparison, we maintain identical core training configurations across all methods, such as the global training batch size and the number of sampled responses per prompt. And all OPCD baselines use the same EMA teacher-update rule as SEPO.
For other hyperparameters, we perform a rigorous sweep, selecting the configuration for each method that yields the highest validation performance across all tasks. 
Comprehensive implementation and hyperparameter settings are provided in Appendix~\ref{app_sec_imple}.


\begin{table*}[t]
\centering
\caption{Performance comparison on science QA and tool use benchmarks, reporting the highest avg@16 achieved within 1 hour and 10 hours of training time. \textbf{Bold} and \underline{underlined} denote the best and second-best performance, respectively. SEPO consistently achieves the optimal or near-optimal performance at both time marks, demonstrating superior training efficiency and performance ceilings.}
\begin{tabular}{lcccccccccc}
\toprule
& \multicolumn{2}{c}{Chemistry} & \multicolumn{2}{c}{Biology} & \multicolumn{2}{c}{Physics} & \multicolumn{2}{c}{Material} & \multicolumn{2}{c}{Tool use} \\
\cmidrule(lr){2-3} \cmidrule(lr){4-5} \cmidrule(lr){6-7} \cmidrule(lr){8-9} \cmidrule(lr){10-11}
& 1h & 10h & 1h & 10h & 1h & 10h & 1h & 10h & 1h & 10h \\
\midrule
\textbf{Qwen3-8B} & \multicolumn{2}{c}{41.5} & \multicolumn{2}{c}{30.5} & \multicolumn{2}{c}{58.6} & \multicolumn{2}{c}{59.8} & \multicolumn{2}{c}{57.1} \\
+ GRPO (off-policy)       & 48.0 & 73.9 & 30.5 & \underline{68.1} & 59.5  & 74.8  & 70.9  & 76.1  & 60.9  & \underline{67.6} \\
+ OPCD (\textit{Self})      & 66.4  & \underline{81.4} & 49.4 & 67.9 & 62.2  & \textbf{77.0} & 67.6  & 74.5  & \underline{63.6}  & 65.7 \\
+ OPCD (\textit{Expert})    & \underline{72.5}  & 78.4 & \underline{50.1} & 57.4    & \textbf{63.6}  & 73.8 & 72.5  & 76.7  & 63.3  & 64.6  \\
+ OPCD (\textit{Label})     & 58.8  & 78.5 & 38.5 & 66.9 & 59.9  & 68.5 & \underline{72.9}  & \underline{77.8}  & 62.5  & 62.5   \\
+ \textbf{SEPO}    & \textbf{73.8}  & \textbf{82.9} & \textbf{55.1} & \textbf{69.0} & \underline{62.8}  & \underline{76.4} & \textbf{73.8}  & \textbf{78.4}   & \textbf{64.6}  & \textbf{68.2}   \\
\midrule
\textbf{Olmo3-7B-Instruct} & \multicolumn{2}{c}{ 25.1} & \multicolumn{2}{c}{13.0 } & \multicolumn{2}{c}{ 37.5 } & \multicolumn{2}{c}{37.6 } & \multicolumn{2}{c}{45.3 } \\
+ GRPO (off-policy)          & 25.4  & 54.7  & 26.0 & 50.6 & 40.5  & 53.2  & 61.1  & 73.9  & 52.7  & \underline{63.4}  \\
+ OPCD (\textit{Self})    & \underline{58.3}  & \underline{80.9}  & 45.1 & 51.5 & 46.3  & \underline{70.2}  & \textbf{65.4}  & \underline{76.9}  & 52.7  & 61.5  \\
+ OPCD (\textit{Expert})  & 43.5  & 64.7  & \underline{45.9} & \underline{56.0} & \textbf{48.4}  & 64.3  & 63.9  & 76.7  & \underline{52.9}  & 54.6  \\
+ OPCD (\textit{Label})   & 34.2  & 54.3  & 30.5 & 46.5 & 38.7  & 49.1  & 62.3  & 66.2  & 51.0  & 60.0  \\
+ \textbf{SEPO}          & \textbf{59.7}  & \textbf{82.1}  & \textbf{46.4} & \textbf{57.4} & \underline{47.0}  & \textbf{71.1}  & \underline{64.7}  & \textbf{78.4}  & \textbf{53.6}  & \textbf{63.8}  \\
\bottomrule
\end{tabular}
\label{tab_main}
\end{table*}


\subsection{Main results}\label{subsec_main_results}
\begin{figure*}[t]
\centering
\includegraphics[scale = 0.48]{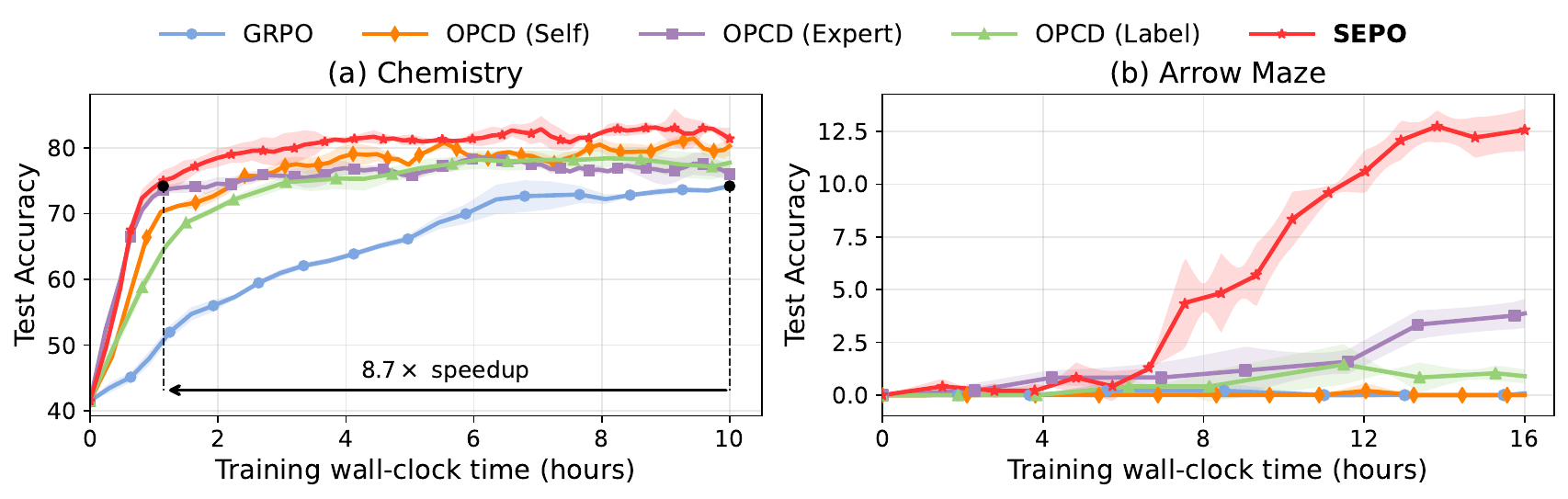}
\caption{Training process on (a) Chemistry and (b) Arrow Maze tasks using Qwen3-8B. Shaded regions show the standard deviation across 3 runs. SEPO consistently maintains top performance throughout training, achieving effective improvements even on Arrow Maze where baselines fail.}
\vspace{-0.2cm}
\label{fig_1_2task}
\end{figure*}

Table~\ref{tab_main} presents a comparison of the peak performance achieved by different methods under a short time budget (1 hour) and at convergence (10 hours) across the scientific and tool-use reasoning domains. 
Figure~\ref{fig_1_2task} (Left) illustrates the training progression of these methods on Chemistry task using Qwen3-8B.
As observed, SEPO consistently secures optimal or near-optimal performance, demonstrating both superior short-term training efficiency and a higher long-term convergence ceiling.

Analyzing the 1-hour results, GRPO consistently underperforms, confirming that OPCD's denser supervision accelerates initial training.
Among OPCD baselines, the \textit{Expert} method achieves rapid early gains, often reaching near-optimal performance at the 1-hour mark, but its 10-hour performance prematurely plateaus.
This indicates that while external demonstrations can bootstrap rapid initial learning, the severe off-policy and distribution mismatch issues impose a strict training ceiling.
Furthermore, merely providing the ground-truth label in the context (\textit{Label}) yields poor final performance on tasks such as Physics and Tool use, where performance either stagnates or degrades after a few steps, suggesting its training instability.
Moreover, while \textit{Self} and GRPO reach comparable convergence levels, their final performance consistently lags behind SEPO, indicating their exploration bottlenecks prevent them from escaping local optima for further gains.
Consequently, by more effectively expanding the exploration boundaries and exploiting the explored paths, SEPO achieves more continuous self-improvement and superior final performance.
Additional experimental results, including on-policy GRPO, standard deviations, and fixed training steps, are in Appendix~\ref{app_sec_more_exp}.

\subsection{SEPO overcomes severe exploration bottlenecks}\label{subsec_arrow_maze}
Having demonstrated SEPO's effectiveness on general reasoning tasks, Figure~\ref{fig_1_2task}(b) further illustrates the 16-hour training processes on Arrow Maze, a highly complex logical reasoning task.
In this significantly harder setting, where the initial pass@16 of Qwen3-8B is strictly zero, both GRPO and \textit{Self} fail entirely. Their inability to overcome the exploration bottleneck prevents them from even initiating effective training.
Although the \textit{Label} and \textit{Expert} methods manage to trigger training, their performance gains are marginal and plateau after only a few steps.
In stark contrast, SEPO exhibits substantial and continuous performance improvements. 
Its guided exploration successfully resolves the zero-accuracy cold-start problem. By coupling this with the effective exploitation of the explored correct reasoning, SEPO forms a robust self-evolution cycle that stably enhances the model's capabilities.
In Appendix~\ref{app_subsec_dyck}, we provide results on another challenging SynLogic task where Qwen3-8B's initial pass@16 is below 20. Although all baseline methods are able to train normally in this regime, SEPO consistently achieves the best performance.



\subsection{SEPO achieves scale-invariant self-improvement}
\begin{wrapfigure}{r}{0.48\textwidth} 
    \centering
    \vspace{-0.5cm}
    \includegraphics[scale=0.45]{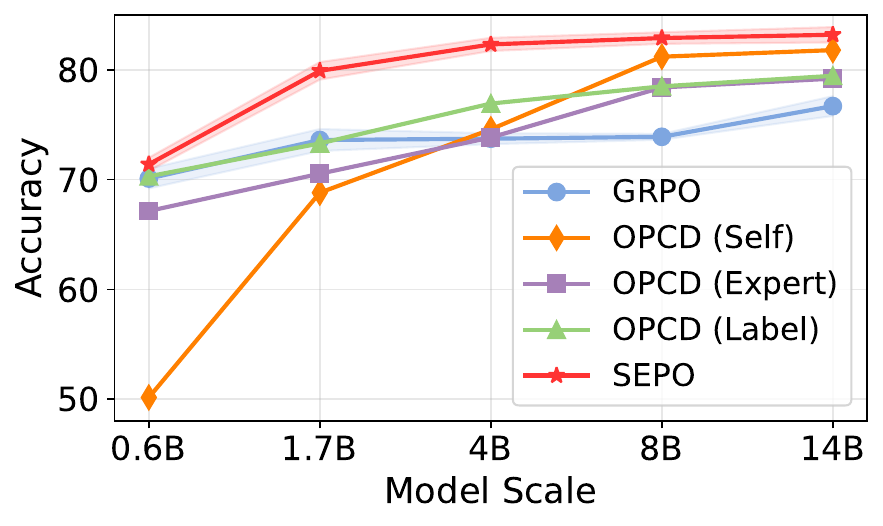}
    \caption{Highest avg@16 on Chemistry after 10h training across Qwen3 scales. SEPO consistently performs best, uniquely outperforming GRPO at 0.6B and 1.7B.}
    \label{fig_scale}
\end{wrapfigure}
Figure~\ref{fig_scale} presents the peak performance achieved by different methods on the Chemistry task after 10 hours of training, evaluated across Qwen3 model scales ranging from 0.6B to 14B.
As observed, SEPO consistently achieves the best performance across all scales.
Comparing OPCD baselines against GRPO, a prominent trend emerges from 0.6B to 8B scales: the performance advantage of OPCD widens as the model scale increases. This suggests that the effectiveness of OPCD methods relies on the policy's intrinsic capacity.
When scaling down to smaller models (e.g., 0.6B and 1.7B), all OPCD baselines underperform GRPO, indicating models with limited in-context integration capabilities struggle to effectively exploit meaningful signals from self-generated reasoning, heavily distribution-shifted demonstrations or sparse labels.
Furthermore, the \textit{Self} method depends not only on model's in-context ability but also on its reasoning exploration capability, resulting in its severely poor performance on small-scale models.
In contrast, SEPO maintains a significant performance advantage over GRPO across different model sizes. This highlights a critical synergy in SEPO's design: by resolving the exploration bottleneck, it constructs congruent in-distribution targets with a low assimilation threshold. This enables even lightweight models to smoothly internalize correct reasoning patterns and achieve continuous self-improvement.

\section{Analysis}


\subsection{SEPO compresses reasoning and maintains exploration diversity}
\begin{figure*}[t]
\centering
\includegraphics[scale = 0.48]{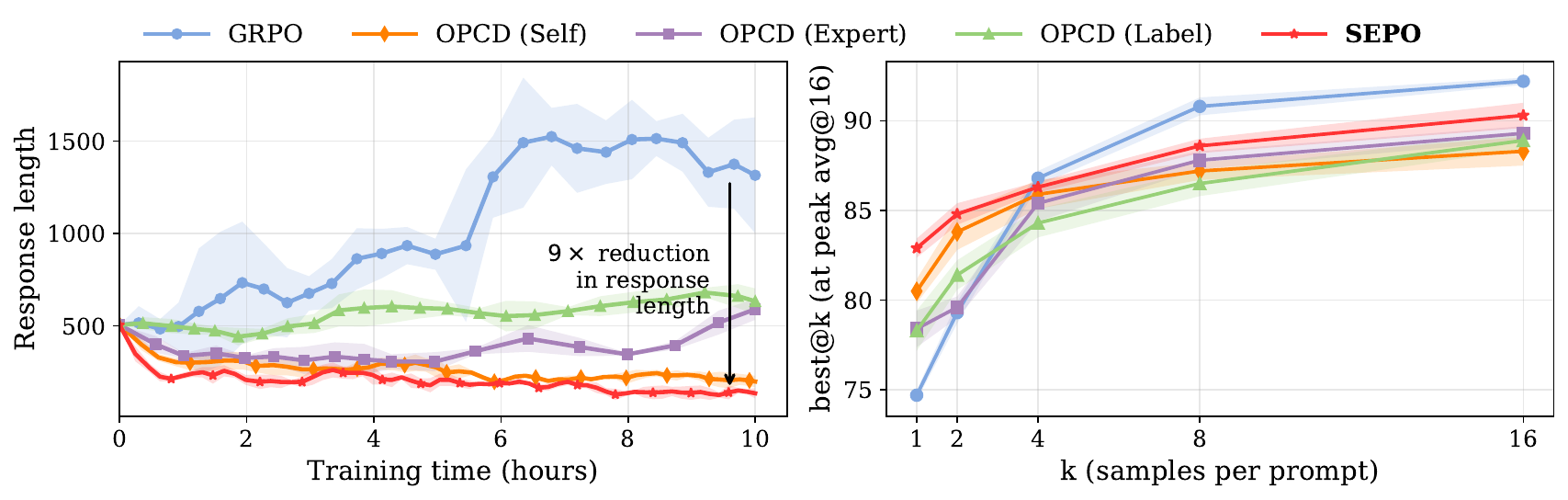}
\vspace{-0.2cm}
\caption{(\textbf{Left}) Response length progression. (\textbf{Right}) Best@$k$ across different $k$ at the highest avg@16 checkpoint. Both evaluations use Qwen3-8B on Chemistry task. SEPO compresses reasoning while achieving superior best@$k$, maintaining exploration diversity.}
\vspace{-0.2cm}
\label{fig_3_ana}
\end{figure*}

Figure~\ref{fig_3_ana} (Left) illustrates the progression of response lengths on Chemistry using Qwen3-8B. While GRPO's lengths increase continuously, OPCD methods maintain or even reduce them.
At convergence, SEPO compresses reasoning to be, on average, more than $5\times$ shorter than those of GRPO.
Although prevailing RLVR methods often rely on scaling solution lengths to unlock advanced capabilities~\cite{DBLP:journals/corr/abs-2512-02556,DBLP:conf/emnlp/MuennighoffYSLFHZLCH25}, our results establish that highly concise reasoning can also remain remarkably effective.

To investigate whether this reasoning compression restricts exploration diversity, Figure~\ref{fig_3_ana} (Right) plots the best@k performance at the peak avg@16 checkpoint. Here, avg@16 (mathematically equivalent to expected best@1) reflects model's expected abilities, whereas best@16 indicates the capability ceiling driven by exploration diversity.
As observed, when k $\geq$ 4, all OPCD methods underperform GRPO, revealing a limited sample variance. 
However, this trade-off is highly favorable for SEPO, which sacrifices merely 2 points in best@16 to secure a massive 10-point advantage in avg@16.
Furthermore, SEPO consistently achieves the highest best@k among OPCD methods across all $k$ values, demonstrating its successful preservation of crucial exploration capabilities despite its shortened reasoning.
By retaining this diverse solution space, SEPO ensures that each training step continues to yield rich and effective supervision signals, which fundamentally drives its accelerated training efficiency and superior convergence performance.

\subsection{SEPO cultivates self-reflective reasoning}
\begin{wrapfigure}{r}{0.45\textwidth} 
    \centering
    \vspace*{-1.5cm}
    \includegraphics[scale=0.63]{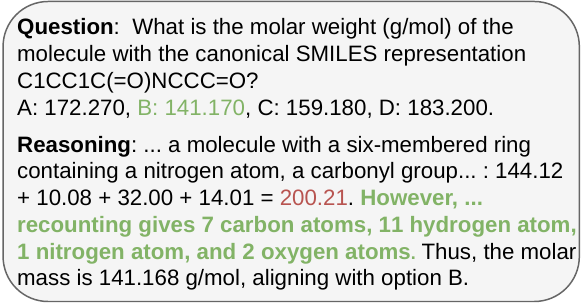}
    \caption{An exemplary reasoning process of SEPO on a Chemistry case, showing a natural self-reflective reasoning pattern.}
    \vspace*{-0.3cm}
    \label{fig_case_study}
\end{wrapfigure}
Building on our quantitative observations of reasoning lengths, we qualitatively analyze the underlying reasoning processes of different methods.
Appendix Table~\ref{app:tab_case_chemistry} presents a typical Chemistry case involving molecular weight computation, and Figure~\ref{fig_case_study} visualizes the simplified SEPO's reasoning.
All methods employ a common forward-deduction strategy, directly calculating the answer and matching it with the options.
However, they diverge significantly in execution.
GRPO exhibits inconsistent atom counting, while \textit{Self} and \textit{Expert} derive incorrect molecular formulas. Moreover, GRPO and \textit{Expert} select answers inconsistent with their own calculations, whereas \textit{Self} obtains an incorrect molecular formula whose calculated mass happens to match a distractor. The \textit{Label} baseline attempts to reconsider its answer but repeatedly performs the same incomplete calculation without correcting the error.
In contrast, SEPO initially misinterprets the molecular structure but subsequently revisits the SMILES representation, correctly identifies the cyclopropane ring, and revises its atom counts to obtain the correct molecular formula and answer. Notably, the label-guided intrinsic teacher reasoning can also contain meaningful corrective steps, as illustrated in Table~\ref{tab:intrinsic_context_hba}. Together, these observations suggest that distilling reasoning-rich teacher contexts may help the student internalize self-correction behaviors, beyond merely learning to match final answers.


\subsection{Ablation study}
\vspace{-0.1cm}
\begin{wrapfigure}{r}{0.42\textwidth} 
    \centering
    \vspace*{-1.8cm}
    \includegraphics[scale=0.41]{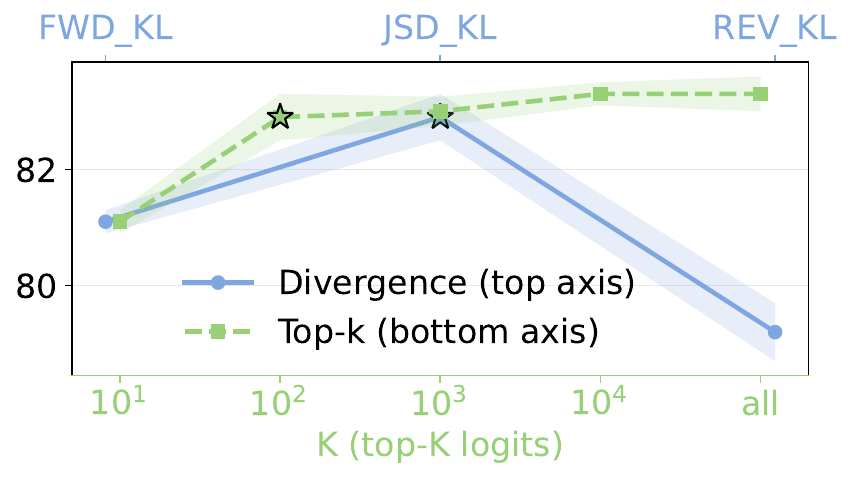}
    \caption{Impact of loss divergence and top-$K$ truncation on 10-hour performance. JSD performs best and top-100 matches full-distribution distillation.}
    \vspace*{-0.4cm}
    \label{fig_abla_dis_obj}
\end{wrapfigure}
\paragraph{Ablation on distillation objectives.}
SEPO introduces critical adjustments to the standard on-policy distillation (OPD) algorithm~\cite{DBLP:conf/iclr/Gu0WH24,lu2025onpolicy}, specifically adopting the JSD objective and restricting the divergence computation to the top-$K$ tokens. Figure~\ref{fig_abla_dis_obj} presents the 10-hour performance of Qwen3-8B on the Chemistry task across various divergence metrics and $K$ values. 
For divergence selection, JSD yields the highest performance, demonstrating the advantage of a balanced, symmetric distillation. 
Furthermore, while traditional OPD favors the mode-seeking Reverse KL, the mode-covering Forward KL actually performs better in our framework.
Regarding the top-$K$ distillation, we observe that model performance plateaus when $K \ge 100$. 
Since the long tail of vocabulary is largely uninformative, this top-$K$ truncation effectively circumvents the massive memory overhead associated with storing full logits for the teacher and student, while preserving the distillation  efficacy.

\vspace{-0.1cm}
\begin{wrapfigure}{r}{0.45\textwidth} 
    \centering
    \vspace{-0.5cm}
    \includegraphics[scale=0.43]{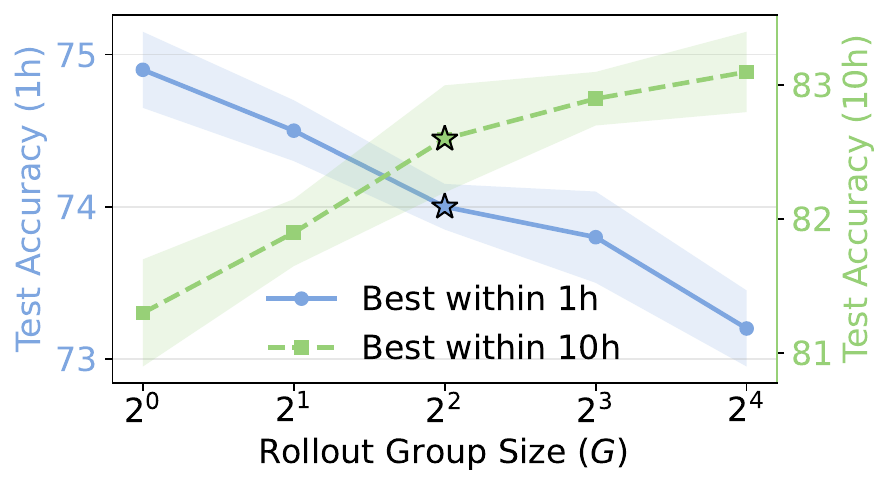}
    \caption{Impact of rollout group size on peak 1-hour and 10-hour performance. Smaller size accelerates training but lower the ceiling, yet any configurations outperform other methods.}
    \vspace{-0.4cm}
    \label{fig_abla_sample}
\end{wrapfigure}
\paragraph{Ablation on rollout group size.}
In Table~\ref{tab_main}, we maintain a consistent rollout group size $G$ per prompt for fair comparisons.
Unlike GRPO and OPCD (\textit{Self}), which rely on sufficiently large rollout pools to discover correct solutions and formulate valid optimization signals, SEPO does not strictly depend on large group sizes. 
Figure~\ref{fig_abla_sample} illustrates the peak performance achieved by Qwen3-8B on the Chemistry task at the 1-hour and 10-hour marks under varying group sizes. A clear trade-off emerges: as the rollout size increases, early 1-hour performance decreases while the final 10-hour performance improves, indicating that short-term training efficiency is traded for a higher convergence ceiling.
Crucially, regardless of the group size, SEPO consistently outperforms all baseline methods at both time scales, underscoring its robustness.

\begin{wraptable}{r}{0.48\textwidth}
\centering
\vspace{-0.45cm}
\setlength{\tabcolsep}{3pt}
\renewcommand{\arraystretch}{1.1}
\caption{Impact of teacher-update strategy on peak 10-hour performance. EMA performs best, while SEPO outperforms OPCD (Self) under both fixed and EMA teachers.}
\label{tab:ema_teacher_ablation}
\begin{tabular}{@{}lrrr@{}}
\hline
\textbf{Method} & \textbf{Chem.} & \textbf{Bio.} & \textbf{Tool Use} \\
\hline
SEPO (Dynamic)    & 47.5 & 45.2 & 61.4 \\
OPCD (Self, Fixed) & 78.5 & 64.3 & 63.7 \\
SEPO (Fixed)       & 80.6 & 65.1 & 66.9 \\
OPCD (Self, EMA)   & 81.4 & 67.9 & 65.7 \\
SEPO (EMA)         & \textbf{82.9} & \textbf{69.0} & \textbf{68.2} \\
\hline
\end{tabular}
\end{wraptable}
\paragraph{Ablation on the teacher-update strategy.}
We evaluate Qwen3-8B with three teacher-update strategies: fully dynamic (synchronized with the current student after each training-batch update), fixed (frozen at the initial policy), and EMA. Within SEPO, we retain label-guided elicitation and distillation while varying the update strategy; fixed teachers still generate label-conditioned context online. We also evaluate OPCD (Self), the strongest baseline under EMA, with a fixed teacher. Table~\ref{tab:ema_teacher_ablation} reports the peak avg@16 achieved within 10 hours on three tasks.
The fully dynamic teacher performs substantially worse, suggesting that directly tracking the current policy creates an unstable distillation target. Fixed-teacher SEPO remains effective, and EMA further improves both SEPO and OPCD (Self), supporting the benefit of gradual teacher adaptation. Crucially, SEPO outperforms OPCD (Self) on all three tasks under both fixed and EMA teachers. Thus, its advantage is not attributable solely to a different teacher-update strategy; these controlled comparisons support label-guided, policy-intrinsic context construction as a key driver of its gains.

\section{Related Work}
\subsection{Reinforcement learning with verifiable rewards}
RLVR methods, such as STaR and GRPO~\cite{zelikman2024star,DBLP:journals/corr/abs-2402-03300}, are widely adopted in LLM post-training to enhance reasoning capabilities~\cite{DBLP:journals/corr/abs-2501-12948,DBLP:journals/corr/abs-2505-09388}. To overcome the challenges of sparse supervision and exploration bottlenecks~\cite{DBLP:journals/corr/abs-2502-01456,DBLP:journals/tmlr/KhalifaALKPLLW26}, recent works leverage external reasoning generated by humans or proprietary models, which fall into two categories. (1) Integrating supervised fine-tuning on expert solutions as an auxiliary loss~\cite{DBLP:journals/corr/abs-2508-11408,DBLP:journals/corr/abs-2506-19767}. (2) Utilizing expert data to guide RL exploration, achieved either by incorporating expert rollouts into the GRPO group for joint learning~\cite{DBLP:journals/corr/abs-2506-19767,DBLP:journals/corr/abs-2504-14945}, or by extracting expert prefixes to serve as initial contexts, facilitating the sampling of correct solutions~\cite{DBLP:journals/corr/abs-2512-17636,DBLP:journals/corr/abs-2601-18779}. 
Unlike these methods that rely on external supervision, SEPO drives simultaneous improvements in training efficiency and final performance entirely through self-evolution distillation.

\subsection{On-policy context distillation}
Recent on-policy context distillation methods drive reasoning improvement by augmenting the current policy's context to construct a stronger teacher for on-policy distillation~\cite{DBLP:journals/corr/HintonVD15,DBLP:conf/iclr/Gu0WH24,lu2025onpolicy,ye2026policy}. 
Various contextual augmentations have been explored: some incorporate environmental feedback, whose applicability is restricted to specific tasks~\cite{DBLP:journals/corr/abs-2601-20802}, while others prompt for simplified teacher responses to compress reasoning~\cite{DBLP:journals/corr/abs-2603-05433}. 
For effective reasoning self-improvement, methods relying on correct student rollouts remain bottlenecked by the model's innate exploration limits, similar to GRPO~\cite{DBLP:journals/corr/abs-2601-20802}. 
Alternatively, injecting external expert reasoning into context suffers from insufficient exploitation of the off-policy solutions~\cite{DBLP:journals/corr/abs-2601-19897,DBLP:journals/corr/abs-2601-18734, DBLP:journals/corr/abs-2512-02556}.
In contrast, the self-explored, in-distribution reasoning within SEPO's context overcomes both the exploration bottleneck and mismatched-distribution exploitation, simultaneously boosting training efficiency and final performance.

\section{Conclusion}
In this work, we develop Self-Evolution Policy Optimization (SEPO), an on-policy self-distillation framework that drives continuous reasoning evolution through superior exploration and exploitation.
By using ground-truth labels as privileged hints, SEPO elicits correct yet strictly in-distribution reasoning from the policy and distills these context-enriched predictions back into the policy. 
Theoretically, we validate that SEPO mitigates both exploration failures and insufficient exploitation issues, guaranteeing sustained capability evolution.
Empirically, experiments across diverse model architectures, scales, and reasoning benchmarks demonstrate SEPO's superior training efficiency and final performance.
Notably, in challenging scenarios where existing baselines fail entirely, such as complex logical tasks or lightweight models, SEPO robustly secures continuous improvements.

\section*{Acknowledgements}
The work is supported in part by the National Natural Science Foundation of China (NSFC) under Grant 62441230, 62502522, 62072458, 62472429 and 62461146205, and in part by the Outstanding Innovative Talents Cultivation Funded Programs 2024 of Renmin University of China.

\bibliographystyle{plain}
\bibliography{neurips_2026}

\newpage

\appendix

\section{Theoretical Analysis}\label{app:sec_theory}

\subsection{Equivalence Between RLVR and OPCD}\label{app:subsec_variation}
\textbf{The Variational Perspective of RLVR}
Standard RLVR methods optimize a policy $\pi_\theta$ to maximize the expected reward while penalizing the deviation from a reference policy $\pi_{\text{ref}}$. The regularized RL objective can be formulated as:
$$\mathcal{J}(\theta) = \mathbb{E}_{x \sim \mathcal{D}} \left[ \mathbb{E}_{\tau \sim \pi_\theta(\cdot \mid x)} \left[ r(\tau) \right] - \eta \mathbb{D}_{\text{KL}} \left( \pi_\theta(\cdot \mid x) \,\|\, \pi_{\text{ref}}(\cdot \mid x) \right) \right].$$
where $\eta > 0$ is a temperature parameter controlling the KL penalty. By taking the variational derivative of $\mathcal{J}(\theta)$ with respect to $\pi_\theta$ and setting it to zero, the closed-form optimal policy $\pi^*$ that maximizes this objective follows an energy-based distribution:$$\pi^*(\tau \mid x) = \frac{1}{Z(x)} \pi_{\text{ref}}(\tau \mid x) \exp\left( \frac{r(\tau)}{\eta} \right),$$where $Z(x) = \sum_{\tau} \pi_{\text{ref}}(\tau \mid x) \exp(r(\tau)/\eta)$ is the intractable partition function~\cite{DBLP:conf/emnlp/KorbakPB22,rafailov2023direct}. Substituting $\pi^*$ back into the objective reveals that maximizing $\mathcal{J}(\theta)$ is mathematically equivalent to minimizing the reverse KL divergence between the current policy and the optimal policy:$$\mathcal{J}(\theta) = \eta \log Z(x) - \eta \mathbb{D}_{\text{KL}} \left( \pi_\theta(\cdot \mid x) \,\|\, \pi^*(\cdot \mid x) \right).$$

\textbf{Equivalence to the Conditional Success Distribution}
Crucially, the partition function $Z(x)$ depends only on the reference policy $\pi_{\text{ref}}$ and the reward environment, meaning it is entirely independent of the active policy parameters $\theta$. Consequently, the term $\eta \log Z(x)$ is a constant with respect to optimization. Maximizing the RL objective is mathematically equivalent to minimizing the trajectory-level divergence between the student and the optimal policy:$$\arg\max_\theta \mathcal{J}(\theta) \equiv \arg\min_\theta \mathbb{D}_{\text{KL}} \left( \pi_\theta(\cdot \mid x) \,\|\, \pi^*(\cdot \mid x) \right).$$
To establish the probabilistic interpretation of $\pi^*$ in the context of logical reasoning, we specialize the environment to binary outcome rewards ($r \in \{0, 1\}$). Under binary rewards, the exponential term $\exp(r(\tau)/\eta)$ acts as a strict filtering mechanism. As $\eta \to 0^+$, the probability mass of incorrect trajectories decays to zero. Consequently, $\pi^*$ mathematically converges to the normalized reference policy restricted to the domain of successful trajectories, which equates precisely to the conditional distribution (as grounded in the framework of control as inference~\cite{DBLP:journals/corr/abs-1805-00909}):$$\pi^*(\tau \mid x) \approx \pi_{\text{ref}}(\tau \mid x \mid \text{Success}).$$Thus, RLVR's true underlying objective is to pull the model toward its own intrinsic distribution of successful reasoning paths.

\textbf{OPCD as Direct Proxy}
Rather than relying on sparse rewards to iteratively approach $\pi^*$, OPCD explicitly constructs a teacher policy $\pi_{\text{tea}}(\cdot \mid x, c)$ using a privileged context $c$ to simulate this conditional success distribution ($\pi_{\text{tea}} \approx \pi^*$). By substituting $\pi_{\text{tea}}$ for $\pi^*$, OPCD minimizes their divergence ($\mathbb{D}_{\text{KL}}(\pi_{\text{tea}} \,\|\, \pi_\theta)$) via dense, token-level supervision. However, the validity of this proxy optimization hinges entirely on $c$ faithfully approximating the natural conditional success distribution, $\pi_{\text{ref}}(\tau \mid x \mid \text{Success})$. If the chosen context induces exploration unavailability, severe distribution shifts from external experts, or non-causal reasoning shortcuts, the theoretical equivalence breaks down, rendering the distilled target invalid.

\subsection{Theoretical Limitations of OPCD Baselines}\label{app:subsec_theory_limit_opcd}

\textbf{The Exploration Bottleneck of Self-Generated Contexts}
When utilizing the student's own correct rollouts as the privileged context ($c = \tau_{\text{self}}$), the teacher distribution implicitly models the conditional success distribution. 
For a given problem $x$, the ability to construct $\pi_{\text{tea}}$ depends on the marginal probability of sampling a successful solution from the current policy $\pi_\theta$:$$P_{\text{self}}(x) = \int \pi_\theta(\tau \mid x) \mathbb{I}(r(\tau) = 1) d\tau.$$In practice, this is approximated by sampling $G$ rollouts. If the task is sufficiently complex or the initial model is weak, the empirical probability $\hat{P}_{\text{self}}(x) = \frac{1}{G} \sum_{i=1}^G \mathbb{I}(r(\tau_i) = 1)$ frequently evaluates to exactly $0$. Mathematically, conditioning on an empty set $\emptyset$ renders the conditional probability $\pi_{\text{tea}}(\cdot \mid x, \tau_{\text{correct}})$ undefined. Consequently, the distillation gradient $\nabla_\theta \mathcal{L}_{\mathrm{CD}}(\theta)$ becomes exactly zero for these problems, completely depriving the model of learning signals and resulting in a severe exploration bottleneck.



\textbf{Static Ceilings and Exploitation Limitations of Expert Contexts}
To bypass exploration failures, using expert demonstrations ($c = \tau_{\text{expert}} \sim \pi_{\text{expert}}$) guarantees context availability. However, since these off-policy contexts are static, they cannot co-evolve with the student's optimization trajectory, which inherently imposes a performance ceiling on continuous self-improvement. 
Even worse, the out-of-distribution (OOD) nature of expert contexts prevents the model from fully exploiting these demonstrations, causing the actual performance to fall short of the theoretical ceiling.
Let the token-level distillation objective be defined by a divergence $\mathbb{D}$. Because the expert possesses a fundamentally different latent reasoning distribution, their support sets are frequently disjoint ($\text{supp}(\pi_{\text{tea}}) \not\subseteq \text{supp}(\pi_\theta)$)~\cite{kumar2019stabilizing,levine2020offline}. This mismatch manifests in two extreme pointwise scenarios: expert-exclusive actions $a^*$ where $\pi_{\text{tea}}(a^* \mid s_t) \to 1$ while $\pi_\theta(a^* \mid s_t) \to 0$, and student-intrinsic actions $a$ where $\pi_\theta(a \mid s_t) > 0$ while $\pi_{\text{tea}}(a \mid s_t) \to 0$. When optimizing across these disjoint supports, the model suffers from extreme penalties that distort its natural behavior:
 \begin{itemize}
 \item \textbf{Forward KL (Forced Inclusion):} For expert-exclusive actions $a^*$, the mode-covering penalty $\pi_{\text{tea}}(a^*) \log \frac{\pi_{\text{tea}}(a^*)}{\pi_\theta(a^*)}$ approaches infinity. To strictly minimize this massive penalty, the optimizer aggressively alters parameters to allocate probability mass to $a^*$. However, this merely forces the inclusion of isolated tokens without reconstructing the underlying causal reasoning chain.
 \item \textbf{Reverse KL (Zero-Forcing):} For student-intrinsic actions $a$, the mode-seeking penalty $\pi_\theta(a) \log \frac{\pi_\theta(a)}{\pi_{\text{tea}}(a)}$ also diverges to infinity. The optimizer suppresses the student's natural probability ($\pi_\theta(a) \to 0$),  eradicating the model's intrinsic deductive stepping stones.\end{itemize}
 Ultimately, standard divergence metrics in OOD regions impose extreme point-wise constraints rather than smoothly integrating the expert's trajectory into the student's natural manifold. The student only acquires specific, isolated patterns and fails to completely transition into the expert's coherent deductive mode.
 This severe under-exploitation, compounded by static context, cements a strict performance ceiling.

\subsection{SEPO's Continuous Improvement}\label{app:subsec_theory_SEPO}
In this section, we provide the rigorous theoretical proof for the accuracy lower bound of the student policy during the Self-Evolution Policy Optimization (SEPO) process.

Let $x \sim \mathcal{D}$ be the problem sampled from the dataset, and let $Y^*(x)$ denote the set of all valid reasoning trajectories or correct final answers for $x$. The expected accuracy of a policy $\pi$ is defined as the expected probability mass it assigns to the correct solutions:$\mathcal{A}(\pi) = \mathbb{E}_{x \sim \mathcal{D}} \left[ \sum_{y \in Y^*(x)} \pi(y \mid x) \right].$During the distillation phase at step $n$, the student policy is updated from $\pi_\theta^{n}$ to $\pi_\theta^{n+1}$ by minimizing the Jensen-Shannon Divergence (JSD) against the teacher $\pi_{\text{tea}}^{n}$. We define the residual exploitation error $\epsilon_{\text{JSD}}$ as the expected JSD remaining after the update:$\epsilon_{\text{JSD}} = \mathbb{E}_{x \sim \mathcal{D}} \left[ \mathbb{D}_{\text{JSD}} \left( \pi_{\text{tea}}^{n}(\cdot \mid x) \,\|\, \pi_\theta^{n+1}(\cdot \mid x) \right) \right].$

\textbf{Proposition 1 (Monotonic Improvement Bound).} Assuming the residual exploitation error is $\epsilon_{\text{JSD}}$, the expected accuracy of the updated student policy $\pi_\theta^{n+1}$ is strictly bounded by:$$\mathcal{A}(\pi_\theta^{n+1}) \ge \mathcal{A}(\pi_\theta^{n}) + \Delta^{n} - \mathcal{O}(\sqrt{\epsilon_{\text{JSD}}})$$where $\Delta^{n} = \mathcal{A}(\pi_{\text{tea}}^{n}) - \mathcal{A}(\pi_\theta^{n})$ is the capability advantage (exploration gap) provided by the teacher. The proof proceeds in three steps: 

\textbf{Step 1: Bounding Accuracy Difference via Total Variation Distance}.
For any given problem $x$, the difference in the probability of generating a correct answer between the teacher $\pi_{\text{tea}}^{n}$ and the updated student $\pi_\theta^{n+1}$ is bounded by the Total Variation distance between their respective distributions. By the definition of TV distance, for any event $A$ (including the event of correctness $Y^*(x)$):$$|\pi_{\text{tea}}^{n}(Y^*(x) \mid x) - \pi_\theta^{n+1}(Y^*(x) \mid x)| \le \mathbb{D}_{\text{TV}}(\pi_{\text{tea}}^{n}(\cdot \mid x), \pi_\theta^{n+1}(\cdot \mid x))$$Taking the expectation over the data distribution $x \sim \mathcal{D}$:$$|\mathcal{A}(\pi_{\text{tea}}^{n}) - \mathcal{A}(\pi_\theta^{n+1})| \le \mathbb{E}_{x \sim \mathcal{D}} \left[ \mathbb{D}_{\text{TV}}(\pi_{\text{tea}}^{n}(\cdot \mid x), \pi_\theta^{n+1}(\cdot \mid x)) \right]$$Removing the absolute value yields the lower bound for the student's accuracy:$$\mathcal{A}(\pi_\theta^{n+1}) \ge \mathcal{A}(\pi_{\text{tea}}^{n}) - \mathbb{E}_{x \sim \mathcal{D}} \left[ \mathbb{D}_{\text{TV}}(\pi_{\text{tea}}^{n}(\cdot \mid x), \pi_\theta^{n+1}(\cdot \mid x)) \right]$$

\textbf{Step 2: Relating Total Variation Distance to Jensen-Shannon Divergence}
To connect the empirical distillation objective to the accuracy bound, we utilize the information-theoretic relationship between TV distance and JSD~\cite{DBLP:journals/tit/Lin91} . For any two probability distributions $P$ and $Q$:$$\mathbb{D}_{\text{TV}}(P, Q) \le \sqrt{2 \ln 2 \cdot \mathbb{D}_{\text{JSD}}(P \,\|\, Q)}$$(Note: If the base of the logarithm in JSD is $2$, the constant $2 \ln 2$ is replaced by $2$. For generality, we represent this constant scaling as $\mathcal{O}(\cdot)$).Applying this inequality to our policies for a specific $x$:$$\mathbb{D}_{\text{TV}}(\pi_{\text{tea}}^{n}(\cdot \mid x), \pi_\theta^{n+1}(\cdot \mid x)) \le \sqrt{2 \ln 2 \cdot \mathbb{D}_{\text{JSD}}(\pi_{\text{tea}}^{n}(\cdot \mid x) \,\|\, \pi_\theta^{n+1}(\cdot \mid x))}$$Taking the expectation over $\mathcal{D}$ and applying Jensen's Inequality (since $f(u) = \sqrt{u}$ is a concave function):$$\mathbb{E}_{x \sim \mathcal{D}} [\mathbb{D}_{\text{TV}}] \le \mathbb{E}_{x \sim \mathcal{D}} \left[ \sqrt{2 \ln 2 \cdot \mathbb{D}_{\text{JSD}}} \right] \le \sqrt{2 \ln 2 \cdot \mathbb{E}_{x \sim \mathcal{D}} [\mathbb{D}_{\text{JSD}}]}$$Substituting our definition of $\epsilon_{\text{JSD}}$:$$\mathbb{E}_{x \sim \mathcal{D}} [\mathbb{D}_{\text{TV}}] \le \sqrt{2 \ln 2 \cdot \epsilon_{\text{JSD}}} = \mathcal{O}(\sqrt{\epsilon_{\text{JSD}}})$$

\textbf{Step 3: Synthesizing the Final Recursive Bound.} Substituting the bound from Step 2 into the inequality from Step 1, we obtain the worst-case performance penalty induced by the exploitation error:$$\mathcal{A}(\pi_\theta^{n+1}) \ge \mathcal{A}(\pi_{\text{tea}}^{n}) - \mathcal{O}(\sqrt{\epsilon_{\text{JSD}}})$$By definition, the exploration gap at step $n$ is $\Delta^{n} = \mathcal{A}(\pi_{\text{tea}}^{n}) - \mathcal{A}(\pi_\theta^{n})$, which implies $\mathcal{A}(\pi_{\text{tea}}^{n}) = \mathcal{A}(\pi_\theta^{n}) + \Delta^{n}$. Substituting this into the inequality, we arrive at the final recursive bound:$$\mathcal{A}(\pi_\theta^{n+1}) \ge \mathcal{A}(\pi_\theta^{n}) + \Delta^{n} - \mathcal{O}(\sqrt{\epsilon_{\text{JSD}}})$$

\section{Limitations}\label{app:sec_limitation}

\subsection{Expert Selection in Baselines}
To ensure a rigorous evaluation of the OPCD (\textit{Expert}) baseline, we evaluate three highly capable models: GPT-OSS-120B, Qwen3-235B, and DeepSeek-V3.2, to generate expert reasoning trajectories. We ultimately select DeepSeek-V3.2 as the external expert, as it yields the highest self-distillation performance. In our comparative analysis, OPCD (\textit{Expert}) exhibits rapid initial training efficiency but suffers from a lower final performance ceiling. Nevertheless, we acknowledge that employing an even stronger future expert, or an expert whose generative distribution is more intrinsically aligned with the student model's preferences, might significantly alter the performance dynamics and upper bounds of this baseline.

\subsection{Computational Overhead of the Elicitation Pipeline}
Compared to the standard on-policy training loops of baseline methods, the SEPO pipeline introduces an additional generation step: it requires the teacher to conditionally elicit the reasoning process based on the ground-truth label. Although SEPO ultimately achieves superior overall training efficiency and convergence performance despite this added step, exploring architectural or procedural optimizations to further streamline this pipeline remains an important direction for future work.

\subsection{Evaluation at Larger Model Scales}
Our experiments evaluate models up to 14B parameters. Although SEPO performs favorably across the evaluated scales, whether these advantages persist for larger models, such as 32B or 72B, remains unverified. Evaluating SEPO at these scales is an important direction for future work.

\section{Broader Impacts}\label{app:sec_broader_impact}
Large Language Models (LLMs) are increasingly deployed across diverse industries, where robust in-domain reasoning capabilities are critical. However, conventional RLVR methods frequently suffer from sparse supervision signals and inherent exploration bottlenecks. Our method, SEPO, offers an efficient and effective self-distillation framework to reliably enhance these reasoning capabilities. This can significantly accelerate the development of highly capable, domain-specific AI systems while democratizing access by reducing reliance on expensive external supervision.

Despite these positive impacts, SEPO also presents potential risks. Because it enables autonomous self-improvement, malicious actors could exploit this framework to accelerate model capabilities in harmful or illicit domains (e.g., generating malicious code or synthesizing dangerous chemical compounds). However, this risk is not unique to SEPO; rather, it is a broader systemic challenge inherent to all algorithms designed for generalized capability enhancement.

\section{Implementation Details}\label{app_sec_imple}

\begin{algorithm}[t]
\caption{SEPO with label-conditioned elicitation and outcome-filtered context}
\label{alg:sepo_complete}
\small
\begin{algorithmic}[1]
\Require Initial parameters $\theta_0$; dataset $\mathcal D$; task verifier $R$; training steps $N$; batch size $B$; student samples $G$;
\Statex \hspace{\algorithmicindent} teacher candidates $H$; threshold $r_{\min}$; EMA rate $\beta$; elicitation template $E$; context template $C$
\State $\theta \gets \theta_0$; $\theta_{\mathrm{tea}} \gets \theta_0$
\For{$n=0,\ldots,N-1$}
    \State Sample $\{(x_i,y_i^*)\}_{i=1}^{B}\sim\mathcal D$; set $\mathcal P\gets\varnothing$
    \For{$i=1,\ldots,B$}
        \State Sample $\tau_{ig}\sim\pi_\theta(\cdot\mid x_i)$ for $g=1,\ldots,G$
        \State Sample $\bar{\tau}_{ih}=(\bar{z}_{ih},\hat{y}_{ih}) \sim \pi_{\theta_{\mathrm{tea}}} (\cdot\mid E(x_i,y_i^*))$ for $h=1,\ldots,H$
        \State $\mathcal Z_i \gets \{\bar z_{ih} :R(x_i,\bar\tau_{ih},y_i^*) \ge r_{\min}\}$

        \If{$\mathcal Z_i\neq\varnothing$}
            \For{$g=1,\ldots,G$}
                \State Draw $z_{ig}^{\mathrm{intrinsic}}\sim\operatorname{Uniform}(\mathcal Z_i)$
                \State Add $(x_i,\tau_{ig},z_{ig}^{\mathrm{intrinsic}})$ to $\mathcal P$
            \EndFor
        \EndIf
    \EndFor
    \For{each optimization mini-batch $\mathcal M\subseteq\mathcal P$}
        \For{each $(x,\tau,z)\in\mathcal M$ and valid response position $t$}
            \State $p_t\gets\pi_\theta(\cdot\mid x,\tau_{<t})$
            \State $q_t\gets\operatorname{stopgrad}\!\left(\pi_{\theta_{\mathrm{tea}}}(\cdot\mid C(x,z),\tau_{<t})\right)$
            \State $S_t\gets\operatorname{TopK}(p_t,K)$
            \State $\bar p_t\gets (p_t(v))_{v\in S_t}\mathbin{\Vert}\left(1-\sum_{v\in S_t}p_t(v)\right)$
            \State $\bar q_t\gets (q_t(v))_{v\in S_t}\mathbin{\Vert}\left(1-\sum_{v\in S_t}q_t(v)\right)$
            \State $m_t\gets(\bar p_t+\bar q_t)/2$
            \State $d_t\gets\tfrac12D_{\mathrm{KL}}(\bar q_t\Vert m_t)+\tfrac12D_{\mathrm{KL}}(\bar p_t\Vert m_t)$
        \EndFor
    \State $\displaystyle
        \mathcal{L}_{\mathcal M} \gets
        \frac{1}{|\mathcal M|}
        \sum_{j=1}^{|\mathcal M|}
        \frac{1}{|\mathcal T_j|}
        \sum_{t\in\mathcal T_j}d_{j,t}$
    \State $\theta \gets
        \operatorname{OptimizerStep}
        (\theta,\nabla_\theta\mathcal{L}_{\mathcal M})$
        
    \EndFor
    \If{a student optimizer update was performed}
        \State $\theta_{\mathrm{tea}}\gets(1-\beta)\theta_{\mathrm{tea}}+\beta\theta$
    \EndIf
\EndFor
\State \Return $\pi_\theta$
\end{algorithmic}
\end{algorithm}

\subsection{SEPO's Implementations}
Algorithm~\ref{alg:sepo_complete} expands the high-level training loop in Algorithm~\ref{alg:sepo}. For every question in each batch, the student generates $G=8$ unhinted rollouts, while the EMA teacher generates $H=4$ candidates using a task-specific label-informed prompt. Teacher elicitation is performed for all questions, rather than only when the student fails. We retain candidates whose final answers pass the task verifier with reward at least $r_{\min}=1$. This outcome-based filter does not certify the correctness of intermediate reasoning. If no candidate is retained for a question, all its student rollouts receive zero distillation weight for that batch; no additional resampling is required. Otherwise, each student rollout is independently paired with a uniformly sampled retained candidate from the same question's pool, with replacement.
We use JSD with mixing coefficient $\lambda=0.5$. At each response position, the student selects the top-$K$ token indices ($K=100$), and both distributions are evaluated on this shared support. Each distribution's remaining probability mass is aggregated into one additional tail category. 
To stabilize updates against distribution shifts, we apply a token-level importance sampling correction, clipping the importance-sampling ratios at $2.0$. Finally, the teacher model is updated via an Exponential Moving Average (EMA) of the student's weights with an update rate of $0.05$.

We implement our training pipeline on top of the verl reinforcement learning framework~\cite{DBLP:conf/eurosys/ShengZYWZZPL025}, utilizing PyTorch Fully Sharded Data Parallel (FSDP2) distributed across 8 H20 GPUs. During the rollout phase, which is accelerated by vLLM~\cite{kwon2023efficient}, we sample $G = 8$ student rollouts per prompt at a temperature of $1.0$. To accommodate extended reasoning traces, we set a maximum prompt length of 2048 tokens and a maximum response length of 22,528 tokens. The policy is optimized using AdamW with a peak learning rate of $1 \times 10^{-5}$, zero weight decay, and a linear warmup over the first 10 steps. Each training step processes a global batch of 32 prompts, which is partitioned into mini-batches of 32 for policy optimization.

\subsection{Enhanced GRPO Implementation}
Following~\cite{DBLP:journals/corr/abs-2601-20802}, we implement an enhanced GRPO baseline that integrates several recent algorithmic improvements. These include asymmetric clipping~\cite{DBLP:journals/corr/abs-2503-14476}, the avoidance of biased normalization~\cite{DBLP:journals/corr/abs-2503-20783}, and off-policy data correction~\cite{yao2025your}.

\subsection{Baseline Implementations}
The implementation of OPCD (\textit{Self}) follows the methodology of~\cite{DBLP:journals/corr/abs-2601-20802}. To construct the self-teacher, each rollout is conditioned on a distinct successful rollout randomly selected from the same group.
For OPCD (\textit{Expert}), we adapt the frameworks established in~\cite{DBLP:journals/corr/abs-2601-19897,DBLP:journals/corr/abs-2601-18734}. To construct the expert dataset, we sample 8 reasoning trajectories per problem and randomly retain one successful rollout to serve as the context. We try three highly capable reasoning models to generate these expert trajectories: GPT-OSS-120B, Qwen3-235B, and DeepSeek-V3.2. We ultimately select DeepSeek-V3.2 as our primary external expert, as it yields the highest self-distillation performance.
The OPCD (\textit{Label}) baseline serves as an ablated variant of the SEPO framework, where the label-guided deduction mechanism is explicitly removed (i.e., the teacher model is merely conditioned on the label, but is not prompted to perform guided deduction toward it).
Following an extensive hyperparameter search, the optimal hyperparameters for all OPCD baselines align with those established for SEPO. For both the on-policy and off-policy variants of GRPO, the optimal learning rate is $1 \times 10^{-5}$.
In addition, for a controlled comparison, SEPO and all three OPCD baselines use an EMA-updated teacher initialized from the same initial student weights.

\subsection{Meta-Prompts}
We detail the specific meta-prompts utilized within the SEPO framework, using the Science QA task as a representative example. Table~\ref{appen:tab_student_prompt} presents the standard base prompt used during student rollouts. Table~\ref{appen:tab_teacher_gen_prompt} displays the prompt used to elicit label-conditioned reasoning from the teacher. Table~\ref{appen:tab_distill_prompt} outlines the prompt employed during the teacher distillation phase.

\begin{table*}[t]
\centering
\caption{Base Meta-Prompt utilized for standard student rollouts.}
\begin{tabular}{p{13cm}}
\toprule
Given a question and four options, please select the right answer. Respond in the following format:\newline
<reasoning> ... </reasoning>\newline
<answer> ... </answer>\newline
For the answer, only output the letter corresponding to the correct option (A, B, C, or D), and nothing else. Do not restate the answer text.\newline
For example, if the answer is "A", just output: <answer>A</answer>\newline
\{Question\}\newline
Please reason step by step. \\
\bottomrule
\end{tabular}
\label{appen:tab_student_prompt}
\end{table*}

\begin{table*}[t]
\centering
\caption{Label-Conditioned Reasoning Prompt utilized for teacher exploration.}
\begin{tabular}{p{13cm}}
\toprule
\{Base Meta-Prompt\}\newline \newline
(Correct answer: \{label\}.\newline Reason as you naturally would without this hint — arrive at the correct answer through genuine understanding, not by working backward from it.) \\
\bottomrule
\end{tabular}
\label{appen:tab_teacher_gen_prompt}
\end{table*}

\begin{table*}[t]
\centering
\caption{Contextual Distillation Prompt utilized during the self-distillation phase.}
\begin{tabular}{p{13cm}}
\toprule
\{Base Meta-Prompt\}\newline \newline
Here is a reference solution:\newline
\{solution\}\newline
\newline
Now, solve the problem on your own.\\
\bottomrule
\end{tabular}
\label{appen:tab_distill_prompt}
\end{table*}


\section{Additional Experimental Results}\label{app_sec_more_exp}

\begin{table*}[t]
\centering
\caption{Supplementary performance comparison between the off-policy and on-policy GRPO variants, extending the results from Table~\ref{tab_main}.}
\begin{tabular}{lcccccccccc}
\toprule
& \multicolumn{2}{c}{Chemistry} & \multicolumn{2}{c}{Biology} & \multicolumn{2}{c}{Physics} & \multicolumn{2}{c}{Material} & \multicolumn{2}{c}{Tool use} \\
\cmidrule(lr){2-3} \cmidrule(lr){4-5} \cmidrule(lr){6-7} \cmidrule(lr){8-9} \cmidrule(lr){10-11}
& 1h & 10h & 1h & 10h & 1h & 10h & 1h & 10h & 1h & 10h \\
\midrule
\textbf{Qwen3-8B} & \multicolumn{2}{c}{41.5} & \multicolumn{2}{c}{30.5} & \multicolumn{2}{c}{58.6} & \multicolumn{2}{c}{59.8} & \multicolumn{2}{c}{57.1} \\
+ GRPO (off-policy)       & 48.0 & 73.9 & 30.5 & 68.1 & 59.5  & 74.8  & 70.9  & 76.1  & 60.9  & 67.6 \\
+ GRPO (on-policy) & 42.5  & 70.5 & 32.4 & 58.0 & 60.5  & 70.2  & 69.3  & 78.1 & 59.3  & 66.5 \\
\midrule
\textbf{Olmo3-7B-Instruct} & \multicolumn{2}{c}{ 25.1} & \multicolumn{2}{c}{13.0 } & \multicolumn{2}{c}{ 37.5 } & \multicolumn{2}{c}{37.6 } & \multicolumn{2}{c}{45.3 } \\
+ GRPO (off-policy)          & 25.4  & 54.7  & 26.0 & 50.6 & 40.5  & 53.2  & 61.1  & 73.9  & 52.7  & 63.4  \\
+ GRPO (on-policy)   & 25.9  & 71.8  & 27.5 & 41.6 & 39.5  & 51.6  & 55.1  & 73.8  & 52.1  & 63.4  \\
\bottomrule
\end{tabular}
\label{app:tab_grpo}
\end{table*}
\subsection{Off-Policy vs. On-Policy GRPO}
In our primary evaluations, we report the off-policy variant of GRPO as the standard GRPO baseline. Table~\ref{app:tab_grpo} details a comprehensive performance comparison between the off-policy and on-policy GRPO variants across various scenarios, utilizing identical experimental settings to those in Table~\ref{tab_main}. As observed, the off-policy variant achieves superior performance in the majority of cases.

\begin{table*}[t]
\centering
\caption{Performance comparison on science QA and tool-use benchmarks, reporting the highest avg@16 achieved within 50 and 400 training steps. Bold and underlined values denote the best and second-best performance, respectively.}
\begin{tabular}{lcccccccccc}
\toprule
& \multicolumn{2}{c}{Chemistry} & \multicolumn{2}{c}{Biology} & \multicolumn{2}{c}{Physics} & \multicolumn{2}{c}{Material} & \multicolumn{2}{c}{Tool use} \\
\cmidrule(lr){2-3} \cmidrule(lr){4-5} \cmidrule(lr){6-7} \cmidrule(lr){8-9} \cmidrule(lr){10-11}
& 50 & 400 & 50 & 400 & 50 & 400 & 50 & 400 & 50 & 400 \\
\midrule
\textbf{Qwen3-8B} & \multicolumn{2}{c}{41.5} & \multicolumn{2}{c}{30.5} & \multicolumn{2}{c}{58.6} & \multicolumn{2}{c}{59.8} & \multicolumn{2}{c}{57.1} \\
+ GRPO (off-policy)       & 63.2  & 73.6  & 51.7 & \underline{68.1} & 67.5  & 76.1  & 72.9  & 77.1  & 60.9  & \underline{68.0}  \\
+ OPCD (\textit{Self})      & 76.1  & \underline{81.7} & \underline{55.0} & 67.9 & 68.2  & \textbf{77.0}  & 70.1  & 74.5  & 63.4  & 65.7  \\
+ OPCD (\textit{Expert})    & \underline{76.2}  & 78.3 & 49.2 & 57.4 & \underline{68.3}  & 73.8  & 73.0  & 78.1  & \underline{63.7}  & 64.6  \\
+ OPCD (\textit{Label})     & 74.2  & 78.9 & 48.2 & 66.9 & 67.3  & 68.5  & \textbf{75.7}  & \underline{77.8}  & 62.5  & 62.5  \\
+ \textbf{SEPO}    & \textbf{76.5}  & \textbf{82.9} & \textbf{58.1} & \textbf{69.0} & \textbf{68.8}  & \underline{76.4}  &\underline{73.8}   & \textbf{78.4}  & \textbf{64.4}  & \textbf{68.2}  \\
\midrule
\textbf{Olmo3-7B-Instruct} & \multicolumn{2}{c}{25.1 } & \multicolumn{2}{c}{13.0 } & \multicolumn{2}{c}{ 37.5 } & \multicolumn{2}{c}{37.6 } & \multicolumn{2}{c}{ 45.3} \\
+ GRPO (off-policy)          &  47.6 & 57.2  & 38.9 & 53.6  & 55.9  & 68.3  & 73.7  & \textbf{79.8}  & 57.5 & \underline{63.4}  \\
+ OPCD (\textit{Self})    & \underline{74.0}  & \underline{80.9}  & 50.5 & 50.5 & \underline{66.9}  & \underline{70.2}  & 74.3  & 77.9  & \underline{60.7}  & 61.5  \\
+ OPCD (\textit{Expert})  & 64.7  & 64.7  & \underline{52.0} & \underline{56.9} & 64.3  & 69.8  & \underline{75.1}  & 76.7  & 54.6  & 54.6  \\
+ OPCD (\textit{Label})   & 44.3  & 54.3  & 46.5 & 46.5 & 49.1  & 49.1  & 66.0  & 66.0  & 60.0  & 60.0  \\
+ SEPO           & \textbf{77.4}  & \textbf{82.1}  & \underline{56.0} & \textbf{57.4} & \textbf{70.1}  & \textbf{73.8}  & \textbf{75.7}  & \underline{79.1}  & \textbf{62.8}  & \textbf{63.8}  \\

\bottomrule
\end{tabular}
\label{app:tab_main_step}
\end{table*}

\subsection{Results under Fixed Training Steps}
In Section~\ref{subsec_main_results} and Table~\ref{tab_main}, we report the peak avg@16 performance achieved by various methods under wall-clock time constraints (1 hour and 10 hours). To complement these findings, Table~\ref{app:tab_main_step} details the highest avg@16 performance attained under fixed budgets of 50 and 400 training steps.
As observed, the comparative trends closely mirror those in Table~\ref{tab_main}. SEPO consistently secures the highest performance across both limited and extended training steps, further validating its superior training efficiency and higher final convergence ceiling.

\begin{table*}[t]
\centering
\caption{Standard deviations across three independent runs corresponding to the main results in Table~\ref{tab_main}.}
\begin{tabular}{lcccccccccc}
\toprule
& \multicolumn{2}{c}{Chemistry} & \multicolumn{2}{c}{Biology} & \multicolumn{2}{c}{Physics} & \multicolumn{2}{c}{Material} & \multicolumn{2}{c}{Tool use} \\
\cmidrule(lr){2-3} \cmidrule(lr){4-5} \cmidrule(lr){6-7} \cmidrule(lr){8-9} \cmidrule(lr){10-11}
& 1h & 10h & 1h & 10h & 1h & 10h & 1h & 10h & 1h & 10h \\
\midrule
\textbf{Qwen3-8B} & \multicolumn{2}{c}{--} & \multicolumn{2}{c}{--} & \multicolumn{2}{c}{--} & \multicolumn{2}{c}{--} & \multicolumn{2}{c}{--} \\
+ GRPO (off-policy)       & 0.42  & 0.27  & 1.05 & 0.41 & 0.82  & 0.44  & 0.96  & 1.05  & 0.85  & 0.49  \\
+ OPCD (\textit{Self})      & 0.31  & 0.67  & 0.92 & 0.55 & 1.01  & 0.53  & 0.88  & 0.42  & 0.62  & 0.94  \\
+ OPCD (\textit{Expert})    & 0.56  & 1.02  & 0.45 & 0.88  & 0.94  & 0.47  & 1.12  & 0.55  & 1.08  & 0.51  \\
+ OPCD (\textit{Label})     & 0.48  & 1.13  & 0.68 & 0.62 & 0.55  & 0.92  & 0.76  & 0.39  & 0.93  & 0.45  \\
+ \textbf{SEPO}    & 0.35  & 0.53  & 0.58 & 0.37 & 0.33  & 0.72  & 0.42  & 0.91  & 0.65  & 0.36  \\
\midrule
\textbf{Olmo3-7B-Instruct} & \multicolumn{2}{c}{--} & \multicolumn{2}{c}{--} & \multicolumn{2}{c}{--} & \multicolumn{2}{c}{--} & \multicolumn{2}{c}{--} \\
+ GRPO (off-policy)           & 0.94  & 0.58  & 0.77 & 0.92  & 1.08  & 0.65  & 1.12  & 0.57  & 0.88  & 0.46  \\
+ OPCD (\textit{Self})      & 0.95  & 0.62  & 0.73 & 0.58 & 0.55  & 0.87  & 0.93  & 0.48  & 0.95  & 0.52  \\
+ OPCD (\textit{Expert})    & 0.88  & 0.49  & 0.89 & 0.44 & 0.96  & 0.52  & 0.74  & 1.05  & 1.16  & 0.64  \\
+ OPCD (\textit{Label})     & 1.12  & 0.55  & 1.02 & 0.78  & 0.83  & 0.63  & 1.01  & 0.59  & 0.42  & 0.98  \\
+ \textbf{SEPO}    & 0.52  & 0.37  & 0.41 & 0.83 & 0.48  & 0.45  & 0.39  & 0.51  & 0.57  & 0.68  \\
\bottomrule
\end{tabular}
\label{app:tab_std}
\end{table*}

\subsection{Standard Deviations Across Independent Runs}
In Table~\ref{tab_main}, we report the peak performance of various methods across different scenarios, averaged over three independent runs with different random seeds. To provide a comprehensive view of training stability, Table~\ref{app:tab_std} details the corresponding standard deviations. As observed, the standard deviations for all methods fall consistently within a tight range of 0.2 to 1.2. This indicates that the observed variances primarily reflect natural experimental fluctuations introduced by stochastic trajectory sampling, rather than underlying algorithmic instability. Consequently, this confirms that the performance gains achieved by SEPO are highly robust and represent stable convergence, independent of specific random seed initializations.

\begin{table*}[t]
\centering
\caption{Performance comparison when models are in thinking mode on science tasks, alongside results on the Dyck Language logical reasoning task.}
\begin{tabular}{lcccccc}
\toprule
& \multicolumn{2}{c}{Chemistry (Thinking)} & \multicolumn{2}{c}{Physics(Thinking)} & \multicolumn{2}{c}{Dyck Language}  \\
\cmidrule(lr){2-3} \cmidrule(lr){4-5} \cmidrule(lr){6-7} 
& 1h & 10h & 1h & 10h & 1h & 10h  \\
\midrule
\textbf{Qwen3-8B} & \multicolumn{2}{c}{56.3} & \multicolumn{2}{c}{70.2} & \multicolumn{2}{c}{12.5}  \\
+ GRPO (off-policy)       & 57.3  & 62.5  & 71.3 & 72.6 & 15.0  & 73.8   \\
+ OPCD (\textit{Self})      & 58.3  & \underline{78.3}  & \textbf{72.2} & \underline{76.3} & \underline{19.8}  & \underline{96.7}   \\
+ OPCD (\textit{Expert})    & \underline{59.8}  & 65.8  & 71.4 & 75.7 & 18.8  & 86.9   \\
+ OPCD (\textit{Label})     & 57.2  & 63.4  & 70.5 & 74.3 & 14.3  & 64.4   \\
+ \textbf{SEPO}    & \underline{60.2}  & \textbf{79.1}  & \underline{71.9} & \textbf{79.2} & \textbf{20.3}  & \textbf{98.1}   \\
\bottomrule
\end{tabular}
\label{app:tab_think_dyck}
\end{table*}

\subsection{Thinking Mode}
In Table~\ref{tab_main}, we report the model's performance in the standard Instruct (non-thinking) mode
Table~\ref{app:tab_think_dyck} extends this evaluation to the Chemistry and Physics tasks under a "Thinking Mode" setting using Qwen3-8B.
As observed, SEPO remains highly effective in this extended reasoning regime, consistently achieving the best performance at both the 1-hour and 10-hour training milestones.

\begin{table*}[t]
\centering
\small
\caption{A complete retained intrinsic teacher trace for the Chemistry question.}
\label{tab:intrinsic_context_hba}
\begingroup
\renewcommand{\arraystretch}{1.08}
\begin{tabular}{@{}p{0.97\textwidth}@{}}
\hline
\textbf{Question}\newline
How many hydrogen bond acceptors are there in the molecule with the canonical SMILES representation 'C=CNCC(=O)N'?\newline
A: 2\quad B: 8\quad C: 6\quad D: 9\quad (answer: A=2)
\\ \hline
\textbf{Complete teacher reasoning}\newline
To determine the number of hydrogen bond acceptors in the molecule represented by the SMILES string 'C=CNCC(=O)N', we need to identify all atoms that can accept hydrogen bonds.\newline
Hydrogen bond acceptors are typically atoms that have lone pairs of electrons and are capable of forming hydrogen bonds with hydrogen atoms. These include: Oxygen (O), Nitrogen (N), Fluorine (F), Sulfur (S)\newline
In the given molecule, the atoms of interest are:
1. The nitrogen in the 'C=CN' group: This is a nitrogen atom in a double bond and has a lone pair, making it a hydrogen bond acceptor. 
2. The oxygen in the 'C(=O)N' group: This is an oxygen double-bonded to carbon and has two lone pairs, making it a hydrogen bond acceptor.
3. The nitrogen in the 'N' at the end: This is a terminal nitrogen atom that has a lone pair and acts as a hydrogen bond acceptor.\newline
No other atoms in the molecule are capable of hydrogen bonding, as the other carbon and hydrogen atoms cannot act as hydrogen bond acceptors.
Thus, there are a total of 3 hydrogen bond acceptors in the molecule: 1 nitrogen, 1 oxygen, and 1 nitrogen. \newline
However, the terminal nitrogen (the last N in C(=O)N) is not a hydrogen-bond acceptor in this context. Let's re-identify the atoms: the non-amide nitrogen in the C=CN group has an available lone pair and can act as a hydrogen-bond acceptor; the oxygen in the C(=O)N group is a double-bonded oxygen with lone pairs and can act as a hydrogen bond acceptor; the carbon and hydrogen atoms can not act as hydrogen bond acceptors. Thus, there are a total of 2 hydrogen bond acceptors: one nitrogen and one oxygen.
\\ \hline
\end{tabular}
\endgroup
\end{table*}

\begin{table*}[t]
\centering
\small
\caption{Responses from different methods after 100 training steps to the question: ``What is the molar weight (g/mol) of the molecule with the canonical SMILES representation C1CC1C(=O)NCCC=O? A: 172.270, B: 141.170, C: 159.180, D: 183.200.'' The correct answer is B. \textcolor{customred}{Red} marks erroneous structural claims, calculations, or an incorrect final answer. \textcolor{customblue}{Blue} marks superficial mechanical reasoning in GRPO, and repeated reflection in Label when the derived answer is inconsistent with the options. \textcolor{customgreen}{Green} denotes spontaneous and corrective reflection in SEPO.}
\label{app:tab_case_chemistry}
\begingroup
\renewcommand{\arraystretch}{1.08}
\begin{tabular}{@{}p{0.08\textwidth}p{0.89\textwidth}@{}}
\hline
\textbf{Method} & \textbf{Output} \\ \hline
\textbf{GRPO} &
To calculate the molar mass of a molecule given its SMILES string 'C1CC1C(=O)NCCC=O', we need to determine the molecular formula and sum the atomic masses: \newline
1. Decompose the SMILES string:
- C1CC1: This is a cyclopropane ring (C3H6). - C(=O): This is a carbonyl group (C=O). - N: A single nitrogen atom. - CCC=O: This is an ethyl group with a double bond to oxygen (C2H5O).\newline
2. Break it down into atoms: \textcolor{customblue}{Carbon (C)}: There are 5 carbon atoms in the ring (C1CC1) + 1 from C(=O) + 2 from CCC=O = 8 C atoms.
\textcolor{customblue}{Hydrogen (H)}: There are 6 H atoms in the ring (C1CC1) + 2 H atoms in CCC=O = 8 H atoms. \textcolor{customblue}{Oxygen (O)}: There is 1 O from C(=O) and 1 O from CCC=O = 2 O atoms. \textcolor{customblue}{Nitrogen (N)}: 1 atom.
\newline
3. \textcolor{customblue}{Atomic masses (g/mol)}:- Carbon (C) = 12.011. - Hydrogen (H) = 1.008. - Oxygen (O) = 16.00.- Nitrogen (N) = 14.007.\newline
4. Calculate molar mass:
$(8 \times 12.011) + (8 \times 1.008) + (2 \times 16.00) + (1 \times 14.007) = 96.088 + 8.064 + 32.00 + 14.007 = \textcolor{customred}{150.159}$ g/mol (rounded to 3 decimal places). \textbackslash n Thus, the correct answer is \textcolor{customred}{C: 159.180}.
\\ \hline
\textbf{\textit{Self}} &
The SMILES representation 'C1CC1C(=O)NCCC=O' creates a molecule that contains: \textcolor{customred}{A five-membered ring with two carbon atoms, one nitrogen, and one oxygen atom (C2N1O1)}, A side chain of three carbon atoms, one hydrogen atom, and two oxygen atoms (C3H1O2), A carbonyl group and a nitrogen group, thus summing up the full molecular composition.
After combining these, the molecular formula is correctly determined as \textcolor{customred}{C$_6$H$_{11}$N$_2$O$_3$}, reflecting the accurate count of atoms and their sources. 
Here are the calculations: Carbon (C): 6 × 12.01 = 72.06 g/mol; Hydrogen (H): 11 × 1.008 = 11.09 g/mol; Nitrogen (N): 2 × 14.01 = 28.02 g/mol; Oxygen (O): 3 × 16.00 = 48.00 g/mol. Total molar mass = 72.06 + 11.09 + 28.02 + 48.00 = \textcolor{customred}{159.17 g/mol}.
This matches Option C (159.180 g/mol), which is the only option that closely aligns with the calculated value and corresponds to the correct molecular formula. Hence, Option C is the most accurate answer.
\\ \hline
\textbf{\textit{Expert}} &
The SMILES string 'C1CC1C(=O)NCCC=O' represents a molecule with a \textcolor{customred}{cyclobutane ring} and a side chain. By breaking down the structure, we determine the molecular formula to be \textcolor{customred}{C8H12NO2}. Calculating the molar mass: Carbon: 8 carbons * 12.01 g/mol = 96.08 g/mol, Hydrogen: 12 hydrogens * 1.008 g/mol = 12.096 g/mol, Nitrogen: 1 nitrogen * 14.01 g/mol = 14.01 g/mol, Oxygen: 2 oxygens * 16.00 g/mol = 32.00 g/mol. 
Total molar mass = 96.08 + 12.096 + 14.01 + 32.00 = \textcolor{customred}{154.186 g/mol}.
After careful analysis, the molecular formula calculation results in a molar mass consistent with the option \textcolor{customred}{C: 159.18 g/mol}.
\\ \hline
\textbf{\textit{Label}} &
Let's break down the SMILES string and identify the atoms and bonds in the molecule. - 'C1CC1': This forms a cyclopropane ring with two carbon atoms connected to the ring. - 'C(=O)': This is a carbonyl group, indicating a carbon double-bonded to an oxygen. - 'N': This is a nitrogen atom. - 'CCC=O': This is a chain of three carbon atoms ending in a carbonyl group (C=O).
So the molecule is a cyclopropane ring (C3), with a carbonyl group (C=O), a nitrogen (N), and a chain of three carbons ending in a carbonyl group. Now, we count each atom and calculate the total molecular weight based on standard atomic weights: Carbon: 8 × 12.011 = 96.088 g/mol, Oxygen: 2 × 16.00 = 32.00 g/mol, Nitrogen: 1 × 14.007 = 14.007 g/mol. Total molecular weight: 96.088 + 32.00 + 14.007 = \textcolor{customred}{142.095 g/mol}\newline
\textcolor{customblue}{But wait---this calculation assumes the structure is that simple.} [\ldots] Now, repeating the molecular weight calculation: [\ldots] Total molecular weight: 96.088 + 32.00 + 14.007 = \textcolor{customred}{142.095 g/mol}\newline
\textcolor{customblue}{[Wait---this still doesn't match the correct answer. Let's go back and accurately analyze the SMILES again.] *4 } 
Adding gives: 96.088 + 32.00 + 14.007 = \textcolor{customred}{142.095 g/mol}. So the correct answer is \textcolor{customred}{C}.
\\ \hline
\textbf{SEPO} &
The SMILES string 'C1CC1C(=O)NCCC=O' describes a molecule with a six-membered ring containing a nitrogen atom, a carbonyl group, and a terminal carbonyl group. A careful count of the atoms gives:- 12 carbon atoms: the ring has 5 carbons, and the remaining 7 carbons from the side chain and other groups. - 10 hydrogen atoms: the hydrogens are located on the ring and the substituents. - 1 nitrogen atom: part of the ring structure. - 2 oxygen atoms: one from the carbonyl group and another from the amide group.
Using the standard atomic weights: Carbon: $12 \times 12.01 = 144.12$ g/mol, Hydrogen: $10 \times 1.008 = 10.08$ g/mol, Oxygen: $2 \times 16.00 = 32.00$ g/mol, Nitrogen: $1 \times 14.01 = 14.01$ g/mol.
Adding these: \textcolor{customred}{$144.12 + 10.08 + 32.00 + 14.01 = 200.21$ g/mol}
\textcolor{customgreen}{However, after revisiting the SMILES, C1CC1 is a three-membered cyclopropane ring. Recounting gives 7 carbon atoms, 11 hydrogen atoms, 1 nitrogen atom, and 2 oxygen atoms. Thus, the molar mass is 141.168 g/mol}, aligning with option B.
\\ \hline
\end{tabular}
\endgroup
\end{table*}

\subsection{Results on the Dyck Language Task}\label{app_subsec_dyck}
In Section~\ref{subsec_arrow_maze}, we demonstrated SEPO's unique ability to initiate training on the Arrow Maze task, an extreme scenario where standard exploration completely fails due to a strict zero initial pass rate. To provide a comprehensive evaluation, Table~\ref{app:tab_think_dyck} details the performance on Dyck Language, another highly complex SynLogic task.
Unlike Arrow Maze, Qwen3-8B exhibits a non-zero initial avg@16 (12.5) on the Dyck Language task. This minimal baseline competence allows all standard methods, including GRPO and \textit{Self}, to successfully bypass the exploration bottleneck and train normally. However, despite this alleviated learning threshold, SEPO still demonstrates superior sample efficiency at 1 hour and achieves the highest final convergence at 10 hours. This confirms that SEPO consistently provides higher-quality optimization signals to maximize performance, regardless of the exploration difficulty.

\section{Case Study}

\subsection{Example of Intrinsic Teacher Context}\label{app:intrinsic_context_example}
To illustrate the privileged context $z_{\mathrm{intrinsic}}$, Table~\ref{tab:intrinsic_context_hba} presents a label-conditioned teacher reasoning trace for a Chemistry question about hydrogen-bond acceptors. The teacher initially identifies three candidate acceptors, including both nitrogen atoms and the carbonyl oxygen. Upon reconsideration, it correctly excludes the terminal amide nitrogen, whose lone pair is delocalized into the adjacent carbonyl group, and identifies the non-amide nitrogen and carbonyl oxygen as the two acceptors. This example illustrates that $z_{\mathrm{intrinsic}}$ can provide meaningful intermediate reasoning and explicit corrective steps, offering richer contextual guidance than the ground-truth label alone.

\subsection{Comparison of Student Reasoning Trajectories}
Appendix Table~\ref{app:tab_case_chemistry} compares representative reasoning trajectories generated by different student policies after 100 training steps on a Chemistry problem involving molecular weight calculation. GRPO initially recognizes the cyclopropane ring but subsequently produces inconsistent atom counts, leading to an incorrect answer. \textit{Self} and \textit{Expert} also derive incorrect molecular formulas, with \textit{Self} obtaining a mass that happens to match an incorrect option and \textit{Expert} selecting an option inconsistent with its own calculation. The \textit{Label} baseline attempts to reconsider its answer but repeatedly performs the same incomplete calculation without correcting the underlying error. In contrast, SEPO initially misinterprets the molecular structure as containing a six-membered ring, but subsequently revisits the SMILES representation and correctly identifies the three-membered cyclopropane ring. It then revises the atom counts to obtain the correct molecular formula, C7H11NO2, yielding a molecular weight of 141.168 g/mol and selecting the correct answer. This case illustrates that SEPO can exhibit meaningful structural correction within its reasoning process, rather than merely matching the final answer.





\end{document}